%% file: main.tex
\documentclass[10pt,twocolumn,letterpaper]{article}

\usepackage[pagenumbers]{cvpr}
\usepackage{booktabs,multirow,makecell,xcolor}

\input{preamble}

\definecolor{cvprblue}{rgb}{0.21,0.49,0.74}
\usepackage[pagebackref,breaklinks,colorlinks,allcolors=cvprblue]{hyperref}

\usepackage{colortbl}

\newcommand{\tianfan}[1]{}
\newcommand{\zt}[1]{}
\newcommand{\td}[1]{}

\newcommand{\thedit}{Ours\xspace}
\newcommand{\thevae}{DA-VAE\xspace}

\def\paperID{8173}
\def\confName{CVPR}
\def\confYear{2026}

\newcommand{\blfootnote}[1]{%
  \begingroup
  \renewcommand\thefootnote{}%
  \footnotetext{#1}%
  \addtocounter{footnote}{-1}%
  \endgroup
}

\title{\thevae: Plug-in Latent Compression for Diffusion via Detail Alignment}

\author{
Xin Cai$^{13}$, Zhiyuan You$^{1}$, Zhoutong Zhang$^{2\dagger}$, Tianfan Xue$^{134}$\vspace{2pt}\\
$^1$Multimedia Laboratory, The Chinese University of Hong Kong\\
$^2$Adobe NextCam \: $^3$Shanghai AI Laboratory \: $^4$CPII under InnoHK \\
{\tt\small caixin025@gmail.com, zhiyuanyou@foxmail.com, zhoutongz@adobe.com, tfxue@ie.cuhk.edu.hk}\\
{\small Project page: \href{https://caixin98.github.io/davae}{\texttt{caixin98.github.io/davae}}}\\
}

\begin{document}

\twocolumn[{
\renewcommand\twocolumn[1][]{#1}
\maketitle
\thispagestyle{empty}
\vspace{-20pt}
\begin{center}
    \includegraphics[width=1.0\linewidth]{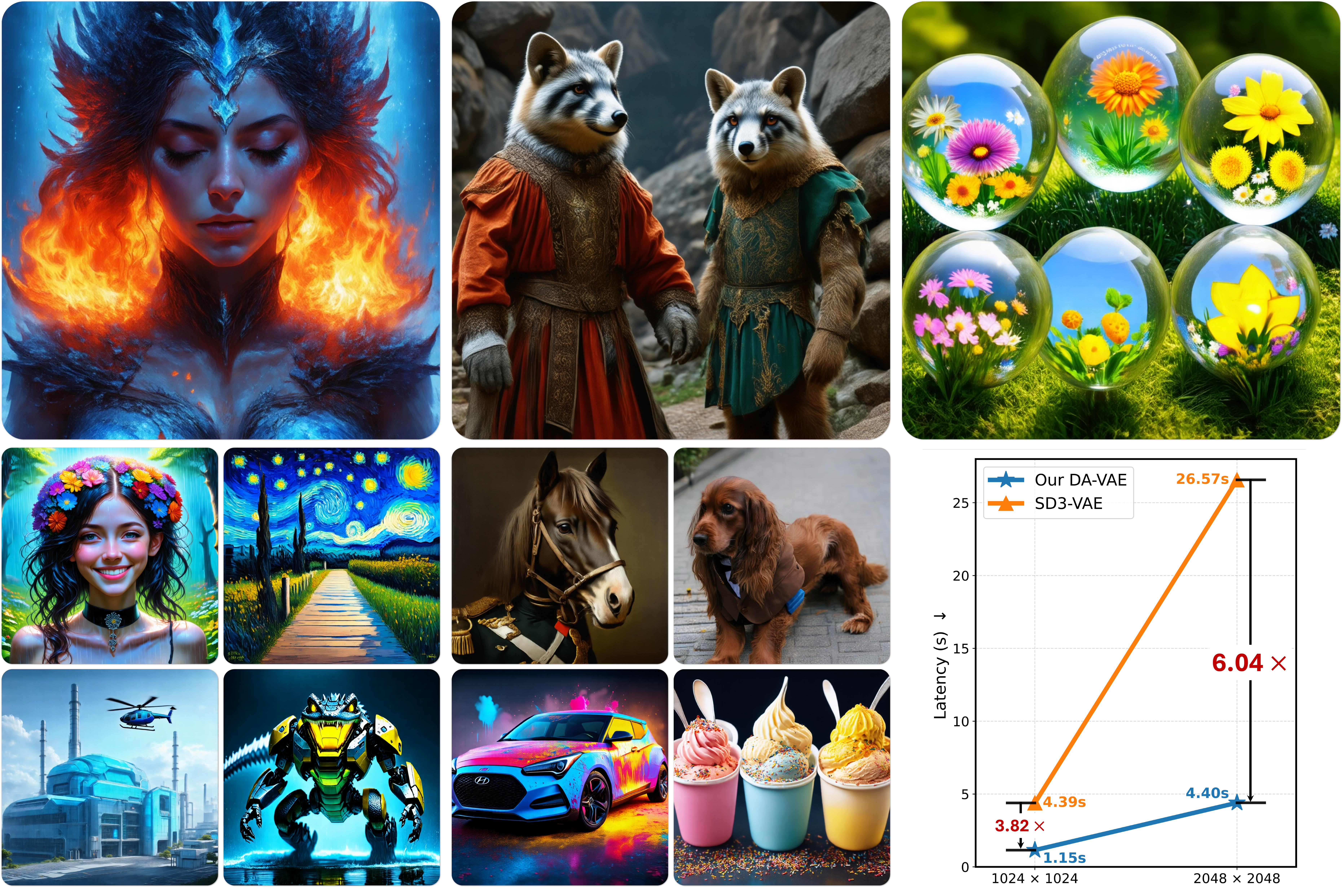}
    \vspace{-20pt}
    \captionof{figure}{
    We propose \textbf{D}etail-\textbf{A}ligned VAE (\textbf{DA-VAE}), a VAE model that increases the compression rate of a pretrained VAE, while requiring only light-weight finetuning of the original diffusion backbone while preserving image quality. Image results are from a finetuned SD3.5 Medium.
    \thevae{} accelerates the original SD3.5 Medium model by 6.04 times for $2048\times2048$ image generation.
    }
\label{fig:teaser}
\end{center}
}]
\blfootnote{$\dagger$ Project lead.}
\input{sec/0_abstract}
\input{sec/1_intro}
\input{sec/2_related}
\input{sec/3_method}

\input{sec/4_exp}

\input{sec/5_final}
{
    \small
    \bibliographystyle{ieeenat_fullname}
    \bibliography{main}
}

\renewcommand\thefigure{S\arabic{figure}}
\renewcommand\thetable{S\arabic{table}}
\renewcommand\theequation{S\arabic{equation}}
\setcounter{section}{0}
\setcounter{equation}{0}
\setcounter{table}{0}
\setcounter{figure}{0}
\renewcommand{\thesection}{S\arabic{section}}

\input{sec/X_suppl}

\end{document}

%% file: preamble.tex
\newcommand{\TODO}[1]{\textbf{\color{red}[TODO: #1]}}

\newcommand{\R}[1]{\noindent \textbf{{\color{darkred}{#1}}}}
\newcommand{\G}[1]{\noindent \textbf{{\color{darkgreen}{#1}}}}
\newcommand{\D}[1]{\noindent \textbf{{\color{darkorange}{#1}}}}

\definecolor{eccvblue}{rgb}{0.12,0.49,0.85}
\definecolor{darkred}{RGB}{180,0,180}
\definecolor{darkgreen}{RGB}{0,130,0}
\definecolor{darkorange}{RGB}{230,120,0}

\usepackage[utf8]{inputenc}
\usepackage{graphicx}
\usepackage{amssymb}
\usepackage{tikz}
\usepackage{makecell}
\usetikzlibrary{positioning,arrows}
\definecolor{mybabyblue}{RGB}{232,244,255}
\tikzset{
  box/.style   ={draw, rounded corners=2pt, inner sep=2.5pt,
                 minimum height=10mm, text width=0.46\columnwidth, align=center},
  latent/.style={box, fill=mybabyblue, draw=blue!50!black},
  iface/.style ={box, fill=blue!6},
  blk/.style   ={box, fill=gray!6}
}
\newlength{\colgap}
\newcommand{\arw}{1pt}
\tikzset{flow/.style={->,thick,shorten >=\arw,shorten <=\arw}}
\usepackage{booktabs}
\usepackage{pifont}
\newcommand{\cmark}{\ding{51}}
\newcommand{\xmark}{\ding{55}}

\renewcommand{\TODO}[1]{}

\renewcommand{\R}[1]{}
\renewcommand{\G}[1]{}
\renewcommand{\D}[1]{}

\usepackage{microtype}

%% file: sec/0_abstract.tex
\begin{abstract}
Reducing the token count is crucial for both efficient training and inference of latent diffusion models, especially at high resolution.
A common approach is to build high-compression-rate image tokenizers that store more information by allocating more channels per token.
However, when trained only with reconstruction objectives, high-dimensional latent spaces often fail to maintain meaningful structure, which in turn complicates diffusion training.
Existing methods introduce additional training targets, such as semantic alignment or selective dropout, to enforce structure in the latent space, but these approaches typically require costly retraining of the diffusion model.
Pretrained diffusion models, however, already exhibit a structured, lower-dimensional latent space; thus, a simpler idea is to expand the latent dimensionality while preserving this structure.
To this end, we propose \textbf{D}etail-\textbf{A}ligned VAE (\thevae{}), a method that increases the compression ratio of a pretrained VAE while requiring only lightweight adaptation for the pretrained diffusion backbone.
Specifically, \thevae{} imposes an explicit latent layout: the first $C$ channels are taken directly from the pretrained VAE at a base resolution, and an additional $D$ channels encode extra details that emerge at higher resolutions.
We introduce a simple detail-alignment mechanism to encourage the expanded latent space to share the structural properties of the original space defined by the first $C$ channels.
Finally, we present a warm-start fine-tuning strategy that enables $1024 \times 1024$ image generation with Stable Diffusion 3.5 using only $32 \times 32$ tokens, $4\times$ fewer than the original model, within a compute budget of 5 H100-days.
It further unlocks $2048 \times 2048$ generation with SD3.5, achieving a $6\times$ speedup while preserving image quality. We also validate the method and its design choices quantitatively on ImageNet.
\end{abstract}

%% file: sec/1_intro.tex
\vspace{-16pt}
\section{Introduction}
\label{sec:intro}
\vspace{-4pt}

Recent text-to-image Diffusion Transformers (DiTs) have achieved state-of-the-art image generation quality.
Various works therefore aim to improve efficiency for those models from different perspectives, such as quantization~\cite{li2024svdquant}, few-step distillation~\cite{yin2024onestep}, and efficient attention mechanisms~\cite{xie2024sana}.
Orthogonal to these aspects, another direction to improve efficiency is through token count reduction.
Since self-attention's computational cost is quadratic over number of tokens, a $4\times$ reduction of tokens would result in $16\times$ reduction for its computational cost.

Existing high-compression-ratio tokenizers \cite{chen2024deep, chen2025dc} are often trained from scratch, aiming to squeeze more pixels into a token with more channels.
Since these works introduce a new latent space, the downstream diffusion model needs to be trained from scratch as well, which requires a tremendous training cost and a large training set.
Adding more to the problem, it is known that high-dimensional features prohibit effective diffusion model training, where one needs to introduce structures to the new latent space, either through semantic alignment or through training time drop-out.
Those challenges combined make this paradigm difficult to iterate—one needs to first train a tokenizer, balance between reconstruction and alignment/auxiliary tasks \cite{yao2025vavae}, then train a generative model from scratch to verify whether the newly introduced latent space is effective for generation.

We propose a different yet simple paradigm to increase the compression ratio of the VAE without complete retraining from scratch.
That is, we start with a pretrained diffusion model, and aim to increase tokenizer efficiency by explicitly introducing a scale-space structure over the channel dimension of each token.
Specifically, with a tokenizer that encodes an image at resolution $H\times W$ with $T$ tokens, we aim to increase the dimension of each token, such that those $T$ tokens can represent an image at a higher resolution of $sH\times sW$.
To achieve this without complete retraining, we keep the first $C$ channels of each token the same as the latent of the image at base resolution.
We then introduce $D$ extra channels to each token, aiming to encode detail information only available at high resolution. With this design, we can then fine-tune the diffusion network trained on original $H \times W$ tokens, as our \thevae ~inherits those tokens.

Still, naively fine-tuning a diffusion denoising network on this new latent space may still fail, since the extra detail channels lack meaningful structure \cite{chen2024deep, chen2025dc}, therefore hindering downstream diffusion training.
To overcome this, we impose an explicit alignment constraint over the detail channels $D$, such that they should have similar structures to the pretrained latent channels $C$.
This alignment proves to be crucial for downstream diffusion training.

Based on this observation, our pipeline is designed as follows.
Since we re-use the original latent in the first $C$ dimension, we use a warm-start strategy to further speed up fine-tuning.
Specifically, we zero initialize the patch embedder for the extra $D$ channels, while using the original patch embedder for the first $C$ channels from the pretrained weights.
We further introduce an optimization schedule that penalizes the extra $D$ channels less during the early training steps. We show that, through ablation studies, this recipe yields better generation results given a fixed training budget.
We further validate our method by fine-tuning SD3.5M \cite{sd35} from $512\times512$ to $1024\times1024$ image generation while keeping the token number fixed.
This results in an overall $\approx4\times$ speedup compared to naive $1024\times1024$ image generation, where adaptation only takes 5 H100-days.
We further demonstrate $2048\times2048$ image generation for SD3.5M with a $6\times$ speedup, while the original model cannot reliably generate coherent structures.

In summary, our contributions are:

\begin{itemize}
    \item A method that improves tokenizer efficiency on the pretrained DiT without costly retraining.
    \item An explicitly structured latent that supports downstream fine-tuning through detail alignment.
    \item A fine-tuning recipe that efficiently adapts a pretrained DiT to the structured latent, enabling $2\times$ higher resolution under the same token budget.
    \item  We validate our method's effectiveness quantitatively on ImageNet and qualitatively by fine-tuning SD3.5, where the adaptation for SD3.5M only takes 5 H100-days.
\end{itemize}

%% file: sec/2_related.tex
\vspace{-7pt}
\section{Related Work}
\label{sec:related}
\vspace{-3pt}

\noindent\textbf{Diffusion model acceleration}.
Diffusion models~\cite{ho2020denoising} achieve strong image quality~\cite{saharia2022photorealistic,esser2024scaling, zhang2025diffusion} but are computationally expensive due to many function evaluations over large latent grids.
Existing acceleration mainly follows three directions. First, improved ODE/SDE solvers and consistency-style objectives reduce the number of sampling steps~\cite{zheng2023dpm,zhao2024unipc,zhangfast,zhanggddim,luo2023latent,zhou2025inductive}.
Second, one- or few-step generators distill the full trajectory into a small number of evaluations~\cite{salimans2022progressive,meng2023distillation,yin2024one,yin2024improved,frans2024one,geng2025mean}.
 Third, per-step efficiency is improved via pruning and token/feature selection, quantization, and optimized attention or execution (e.g., compiler stacks, parallelization, caching)~\cite{fang2024structural,zhao2024vidit,ansel2024pytorch,xie2024sana,xie2025sana,wang2024pipefusion,ma2024deepcache,shih2024parallel,tang2024accelerating}.
 However, these methods preserve the tokenization scheme and thus still scale quadratically with the number of latent tokens. We instead target \emph{token efficiency}: we design a structured latent representation that enables DiTs to operate with substantially fewer tokens while keeping the backbone and sampler largely unchanged, and our tokenizer can be combined with the above techniques for further speedup.

\vspace{2pt}\noindent\textbf{Image tokenizers for generation}.
Latent diffusion models replace pixel-space denoising with denoising in a lower-resolution latent space learned by a 2D continuous tokenizer, typically an $8\times$ VAE~\cite{rombach2022high}. This design is widely adopted by subsequent systems, which mainly scale model size and data while keeping a similar tokenizer~\cite{zhu2023designing,podell2023sdxl,dai2023emu,bao2023all,peebles2023scalable,chenpixart,chen2024pixart,esser2024scaling,flux2024}, so the number of latent tokens still grows quadratically with image resolution.
To reduce this token burden, recent work proposes more aggressively compressed tokenizers: DC-AEs and follow-ups~\cite{chen2024deep,chen2025dc} build deeper encoder--decoder hierarchies that operate at higher spatial downsampling factors, and several 1D tokenizers~\cite{chen2024softvq,chen2025masked,kim2025democratizing,yu2024image, xie2025layton} further lower the number of tokens fed into the diffusion backbone.
However, these token-reduction schemes typically require training a new generative model directly on the new latent space \cite{xie2024sana,xie2025sana}.

A complementary direction focuses less on compression and more on learning \emph{semantically aligned} latent spaces that improve the trade-off between reconstruction and generation \citep{zheng2025diffusion,yao2025vavae,shi2025latent,yue2025uniflow,chen2025aligning}. Prior work shows that shaping the latent geometry to better match semantic structure can yield more robust generation.
However, these methods mainly focus on improving the global semantic structure of the latent space, and pay less attention to preserving structured fine-grained details that are critical for high-resolution image synthesis.

Our method is orthogonal to both deep-compression and semantic-alignment tokenizers. We instead target a \emph{token-efficient} latent space that remains compatible with an existing pretrained DiT: rather than discarding the original tokenizer, we introduce a structured base--detail composition and explicitly align the new latent representation to the original VAE space, allowing us to reduce tokens while keeping the DiT backbone and training objective largely unchanged.

\vspace{2pt}\noindent\textbf{Efficient autoencoder adaptation}.
A related line of work studies how to upgrade or replace the autoencoder while reusing as much of the generative backbone as possible.
Previous work such as \cite{chen2024pixart,peng2025open} adopts a stronger tokenizer and adapts the DiT on top of it for high-resolution generation, but the retraining pipeline is still computationally expensive.
Concurrent work DC-Gen~\cite{he2025dc,chen2025dc} adapts a pretrained DiT to a new, more compressed latent space. However, this target space differs from the original VAE latent space, and the mismatch between them is non-trivial to compensate.
In contrast, our method keeps the original VAE latent space as a reference: we introduce a structured latent space and train a compressor that reduces the number of tokens in an aligned latent space, so that the pretrained DiT can be reused with minimal modification.

%% file: sec/3_method.tex
\vspace{-4pt}
\section{Method}
\input{figures/method}
\vspace{-2pt}

State-of-the-art text-to-image diffusion models~\cite{flux2024, rombach2022high, podell2023sdxl} typically compress latent spaces to reduce computational cost.
To describe the compression ability of a tokenizer, three metrics are often used: feature down-sampling rate $f$, number of channels per token $C$, and the patch size $p$ of the downstream diffusion model.
For an image of size $H\times W$, the latent dimension of all tokens is $(H\times W \times C) / (f^2\times p^2)$.
To increase token efficiency, previous works \cite{xie2024sana} have shown that increasing the downsampling ratio $f$ is both efficient and friendly to downstream generation model training than increasing the patch size $p$.
Moreover, simply increasing the downsampling ratio $f$ limits the reconstruction abilities of the tokenizer, and previous tokenizers  \cite{chen2025dc,chen2024deep} therefore increase the channel number $C$ to compensate for that.

However, increasing $C$ naively brings many challenges to the downstream diffusion training. As discussed in~\cite{yao2025vavae, chen2025dc}, training a diffusion model on wide-channel tokens is unstable, and semantic alignment or auxiliary tasks are often needed to increase convergence of diffusion.
This process usually requires retraining both tokenizer and the diffusion model from scratch, which is prohibitively expensive, introducing significant training cost and data collection cost.

We therefore propose \thevae~ together with a fine-tuning recipe that can both reduce the token size (larger downsampling ratio $f$) and generate friendly tokens for diffusion training.
Core to our method is an explicitly structured latent space with an alignment strategy, and a warm-start diffusion fine-tuning recipe that adapts to the structured latent within a reasonable compute budget.

\subsection{Structured Latent and Alignment}
 We describe the designs and training of \thevae, which increases a pretrained VAE’s spatial compression rate $f$ and number of channels $C$ with a structured latent space.

\noindent\textbf{Structured latent space}.
We start with a pretrained VAE encoder $E$ that encodes an image $\mathbf{I}$ of resolution $H\times W$ into latent space $z$, whose dimensions are $C\times\frac{H}{f}\times\frac{W}{f}$.
To improve its efficiency, we encode a higher resolution image $\mathbf{I}_{hr}$ of size $sH\times sW$ into latent $z_{hr}$ of size $C’\times\frac{H}{f}\times\frac{W}{f}$, where $C’ > C$. In our experiment, we set $s=2$.

Our structured latent is designed such that $C’ = C + D$, where the first $C$ channels are exactly the latent of $\mathbf{I}$. We augment the latent with an additional $D$-channel detail branch, encoded from a separate encoder $E_{hr}$ using $I_{hr}$.

Specifically, our structured latent $z_{hr}$ is composed of two parts concatenated over the channel dimension:
\vspace{-2pt}
\begin{equation}
\mathbf{z_{hr}} \;=\; [\,\mathbf{z} \,,\, \mathbf{z}_d\,]
\;\in\; \mathbb{R}^{(C{+}D) \times \frac{H}{f}\times \frac{W}{f}},
\label{eq:decode1}
\end{equation}
where
\vspace{-2pt}
\begin{equation}
\begin{split}
\mathbf{z} &\;=\; E(I)
\;\in\; \mathbb{R}^{C \times \frac{H}{f}\times \frac{W}{f}}, \\
\mathbf{z_d} &\;=\; E_{d}(I_{hr})
\;\in\; \mathbb{R}^{D \times \frac{H}{f}\times \frac{W}{f}}.
\label{eq:decode2}
\end{split}
\end{equation}
\vspace{-2pt}

To decode the structured latent into image, we use a single decoder $D$ such that $D(z_{hr})$ reconstructs $I_{hr}$.
Throughout our experiments, we keep $E$ fixed to its pretrained weights, and only optimize parameters of $E_{d}$ and $D$. Fig.~\ref{fig:method} shows this design.

\input{figures/alignment_viz}

\noindent\textbf{Latent alignment}.
Naively adding more channels to the existing latent makes diffusion training difficult.
When the VAE encoder is trained only with a reconstruction loss, the extra detail channels $\mathbf{z}_d$ tend to absorb noisy residuals rather than forming a meaningful semantic structure.
As shown in the right column of \cref{fig:latent_align_viz}, features derived from $\mathbf{z}_d$ are poorly organized and weakly correlated with either the original latent $\mathbf{z}$ or the underlying labels, which makes them hard for the downstream diffusion model to exploit.

To regularize these channels, inspired by recent semantic-alignment works, we introduce a latent alignment loss that encourages $\mathbf{z}_d$ to be consistent with the pretrained latent $\mathbf{z}$.
Specifically, we minimize
\vspace{-2pt}
\begin{equation}
\mathcal{L}_{\text{align}}
\;=\;
\big\|\, \mathrm{Proj}(\mathbf{z}_d) - \mathbf{z} \,\big\|^2,
\label{eq:align}
\end{equation}
where $\mathrm{Proj}(\cdot):\mathbb{R}^{D \times H \times W} \to R^{C \times H \times W}$ is a parameter-free channel-wise grouped reduction, defined as:
\vspace{-2pt}
\begin{equation}
\mathrm{Proj}(\mathbf{z}_r)[i, h, w]
\;=\;
\frac{1}{r}\sum_{j=1}^{r} \mathbf{z}_d[ir+j,h,w],
\label{eq:groupavg}
\end{equation}
\vspace{-2pt}where $r = D/C$, and $i,h,w$ are indices over channel, height and width.
Note that for special cases where $r=1$, this reduces to a simple $L_2$ norm between $z_d$ and $z$.

With the alignment loss, the additional latent $\mathbf{z}_d$ learns a structure that closely mirrors $\mathbf{z}$ rather than drifting to arbitrary residuals, as illustrated in the left column of \cref{fig:latent_align_viz}.
This makes the enriched latent representation much more amenable to downstream diffusion training.

\noindent\textbf{Objective for VAE}.
In addition to alignment, we adopt the standard reconstruction losses for VAE training, \ie, perceptual loss, $L1$ loss, adversarial loss and KL regularization:
\vspace{-2pt}
\begin{equation}
\mathcal{L}_{\text{rec}}
\;=\;
\lambda_{\text{L}}\,\mathrm{LPIPS}
\;+\;
\lambda_{1}L_1
\;+\;
\lambda_{\text{adv}}\,\mathcal{L}_{\text{adv}} + \lambda_{KL}L_{KL}.
\label{eq:recon}
\end{equation}
The full training loss is therefore given by:
\begin{equation}
\mathcal{L}
\;=\;
\mathcal{L}_{\text{rec}}
\;+\;
\lambda_{\text{align}}\,
\mathcal{L}_{\text{align}}.
\label{eq:total}
\end{equation}

We empirically show that alignment introduces slight degradation in terms of reconstruction, yet greatly boosts generation, as shown by our experiments in \cref{tab:imagenet-vae}.

\subsection{Warm Start for Diffusion Fine-tuning}
\label{subsec:warm_start}

To adapt the pretrained DiT to the new latent space, we introduce a zero-init strategy and a gradual loss scheduling such that fine-tuning the diffusion model can be warm-started effectively from pretrained weights.

\noindent\textbf{A zero-init strategy}.
As shown in right part of \cref{fig:method}, the Diffusion Transformer uses a patch embedder $P$ to map image latents into the DiT high-dimensional space, i.e.,
\vspace{-2pt}
\begin{equation}
P(\cdot): \mathbb{R}^{C \times \frac{H}{f} \times \frac{W}{f}} \to \mathbb{R}^{L \times \frac{H}{fp} \times \frac{W}{fp}}.
\end{equation}
\vspace{-2pt}At the end of the network, an output layer $O$ maps the DiT features back to the image latent space,
\vspace{-2pt}
\begin{equation}
O(\cdot): \mathbb{R}^{L \times \frac{H}{fp} \times \frac{W}{fp}} \to \mathbb{R}^{C \times \frac{H}{f} \times \frac{W}{f}}.
\end{equation}

To accommodate our new latent space with more channels, we introduce an additional patch embedder $P’$ and output layer $O’$:
\vspace{-2pt}
\begin{equation}
\begin{aligned}
P’(\cdot)&: \mathbb{R}^{D \times \frac{H}{f} \times \frac{W}{f}}
\to \mathbb{R}^{L \times \frac{H}{fp} \times \frac{W}{fp}}, \\
O’(\cdot)&: \mathbb{R}^{L \times \frac{H}{fp} \times \frac{W}{fp}}
\to \mathbb{R}^{D \times \frac{H}{f} \times \frac{W}{f}}.
\end{aligned}
\end{equation}
\vspace{-2pt}
Under this design, the input to the DiT space is
\vspace{-2pt}
\begin{equation}
P(z) + P’(z_{\mathrm{hr}}),
\end{equation}
\vspace{-2pt}
and the DiT output $L$ is decoded to the latent spaces via
\vspace{-2pt}
\begin{equation}
\hat{u} = O(L), \qquad \hat{u}_{\mathrm{hr}} = O’(L),
\end{equation}
\vspace{-2pt}where $\hat{u}$ and $\hat{u}_{\mathrm{hr}}$ are the predictions for the original and high-resolution latents, respectively.

\input{figures/initialization}

To preserve the pretrained behavior, we keep the original latent $\mathbf{z}$ path intact and zero-initialize $P’$ and $O’$ so that their outputs are zero at initialization. In this way, the overall model is exactly equivalent to the pretrained DiT at the beginning of fine-tuning, so the learned priors are fully preserved and training starts from a valid diffusion model. As illustrated in ~\cref{fig:initialization}, this zero initialization leads to a much more stable optimization and significantly faster convergence compared to standard random initialization.

\noindent\textbf{Gradual loss scheduling}.
Besides the zero-init strategy, we introduce a loss scheduling that allows the diffusion fine-tuning process to gradually adapt to the extra channels.
Specifically, we introduce a cosine-annealed weighting strategy applied to the diffusion training loss.
Let $\hat{\boldsymbol{u}}_{hr} = [\hat{u}, \hat{u}_d]$ denote the DiT’s prediction on the structured latent
\([\mathbf{z}_b\,\|\,\mathbf{z}_r]\), and $\boldsymbol{u}_{hr} = [\boldsymbol{u}, \boldsymbol{u}_d]$ are the target of $v$ parameterization.
We introduce a loss scheduling weight as:
\vspace{-2pt}
\begin{equation}
w(n) =
\begin{cases}
\tfrac{1 - \cos(\pi\, n / N_{\mathrm{warm}})}{2}, & n< N_{warm} \\
1, & n\geq N_{warm}
\end{cases}
\end{equation}
\vspace{-2pt}where \(N_{\mathrm{warm}}\) is a hyper-parameter for the number of warm-up steps, and $n$ is the current training step.
We apply this weight only to $\hat{u}_d$, that is:
\vspace{-2pt}
\begin{equation}
\footnotesize
{
\mathcal{L}_{\text{DiT}}(n)
=
\frac{1}{|B| + w(n) |R|}
(
\big\|
\hat{\boldsymbol{u}} - \boldsymbol{u}
\big\|_2^2 +
w(n)\big\|
\hat{\boldsymbol{u}}_{d} - \boldsymbol{u}_{d}
\big\|_2^2)
}
\end{equation}
\vspace{-2pt}At early iterations (\(w(n) \approx 0\)), gradients are dominated by the base latent channels,
ensuring stable alignment with the pretrained backbone.
As training proceeds, the diffusion model is gradually forced to model the extra detail channels \(\mathbf{z}_{d}\).
This scheduling effectively regularizes the pretrained model to gradually adapt to the new latents.

\noindent\textbf{End-to-end fine-tuning}.
With the above recipes,  we fine-tune all blocks of a pretrained DiT, together with both patch embedders $P$, $P’$ and output layers $O$ and $O’$.
For large model SD3.5~\cite{sd35}, we use LoRA on all attention and FFN layers, but still optimize all parameters of $P$, $P’$, $O$ and $O’$.

%% file: figures/method.tex
\begin{figure*}
    \centering
    \includegraphics[width=0.95\linewidth]{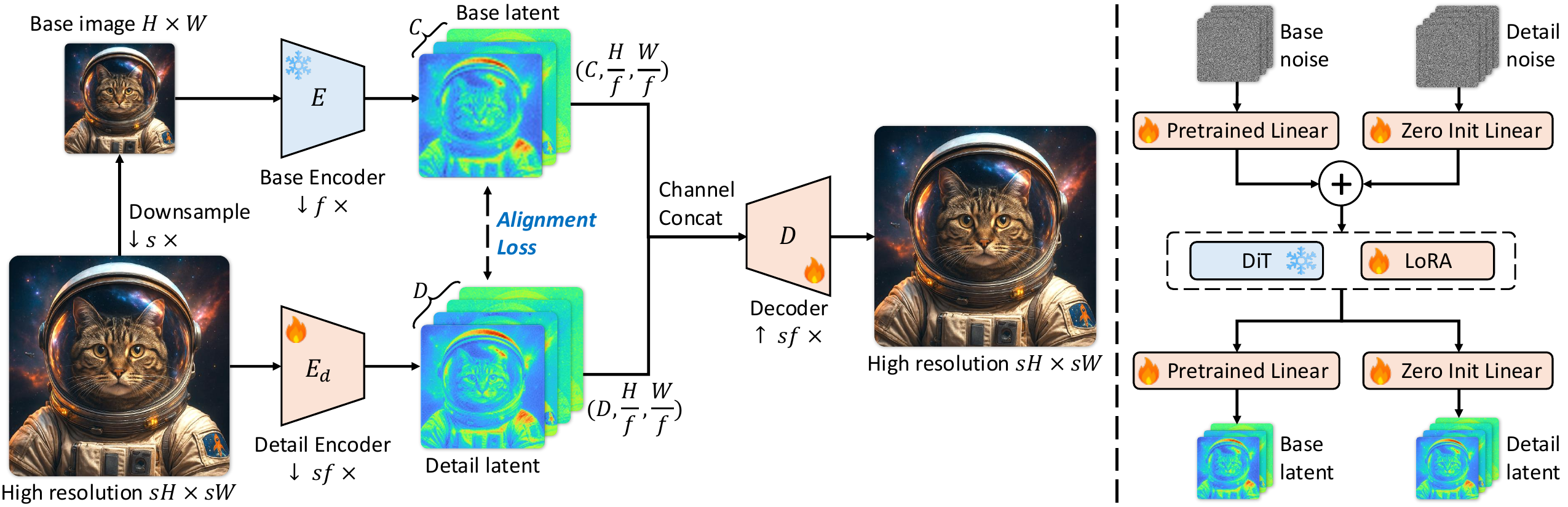}
    \vspace{-9pt}
    \caption{
    \textbf{Overview of our method}. Left: \textbf{Illustration of our Detail-Aligned VAE (DA-VAE)}, which encodes a high-resolution image using the same number of visual tokens as the base image. Right: \textbf{Zero initialization} of the linear layer for detail latent. At the beginning of training, the model keep pretrained diffusion model capability of generating images at the base resolution. 
    }
    \label{fig:method}
    \vspace{-15pt}
\end{figure*}

%% file: figures/alignment_viz.tex
\begin{figure}
    \centering
    \includegraphics[width=1.0\linewidth]{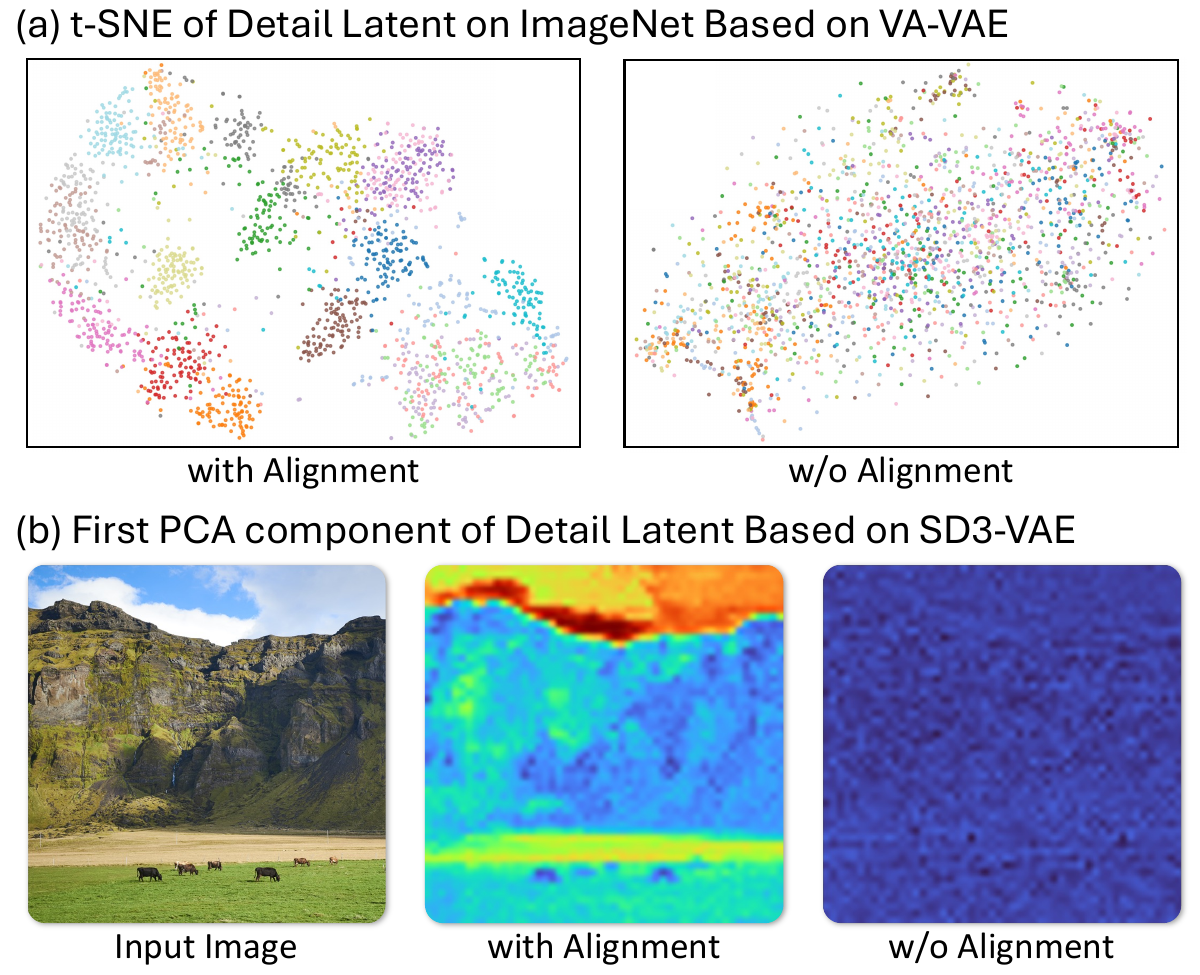}
    \vspace{-20pt}
    \caption{
    \textbf{Effect of the proposed latent alignment loss} on the learned detail feature $\mathbf{z}_d$. 
    In each pair, the right column shows training with only reconstruction loss (no alignment), and the left column shows training with our alignment loss.
    (a) DA-VAE on VA-VAE: the detail features (points colored by class) become more class-separable and well clustered under alignment, suggesting that $\mathbf{z}_d$ inherits the semantic structure of the original latent. 
    (b) DA-VAE on SD3-VAE: alignment encourages the detail branch to capture fine-grained textures while preserving the global image layout, instead of collapsing into noisy residuals.
    }
    \label{fig:latent_align_viz}
    \vspace{-18pt}
\end{figure}

%% file: figures/initialization.tex
\begin{figure}[t]
  \centering
  \includegraphics[width=\linewidth]{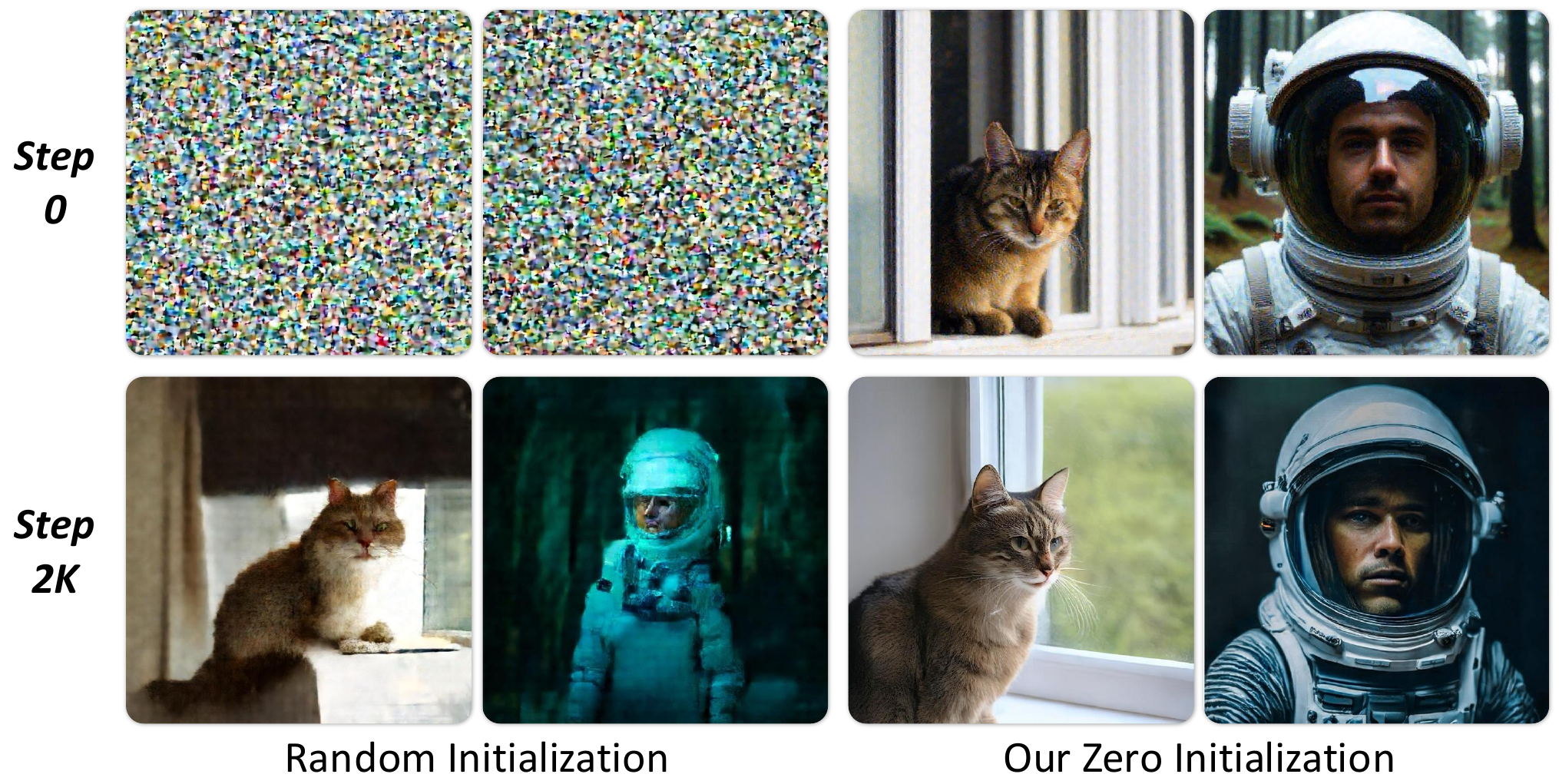}
  \vspace{-20pt}
  \caption{
  \textbf{Comparison of zero initialization and random initialization}. Benefiting from our zero initialization, the model starts from a well-behaved point and converges faster during training.
  }
  \label{fig:initialization}
  \vspace{-18pt}
\end{figure}

%% file: sec/4_exp.tex
\input{tables/imagenet}
\input{figures/res_imagenet}
\input{tables/imagenet-vae}

\vspace{-3pt}
\section{Experiments}
\vspace{-1pt}

We evaluate our method on both ImageNet and general text-to-image generation.
On ImageNet, we show both qualitative and quantitative results by fine-tuning a base model to generate $512\times 512$ images.
To show the importance of individual components of our method, we perform ablation studies with quantitative results.
For general text-to-image experiments, we fine-tune over SD3.5 Medium with LoRA and report qualitative and quantitative results, starting from a base resolution of $512\times 512$ to generate $1024\times 1024$ images.
We also show qualitative results by fine-tuning from $1024\times1024$ to generate $2048\times2048$ images.

\input{figures/compare_sd35}
\input{figures/compare_sd35_2k}

\vspace{2pt}\noindent\textbf{ImageNet experiment details}. 
We use the pretrained VA-VAE and LightningDiT-XL from \cite{yao2025vavae} as our base model, which is trained to generate images at the resolution of $256\times256$.
VA-VAE  uses a spatial compression factor $f=16$ and $C=32$ latent channels.  Our structural latent adds additional $D=96$ channels to the latent space, resulting in a total of 128 channels for each token. 
LightningDiT-XL is a DiT-XL/1~\citep{peebles2023scalable} diffusion model trained on the latent space of VA-VAE for $256 \times 256$ image generation, with a patch size $p=1$.
\thevae~is trained for 100k steps with a batch size of 1024 and the DiT backbone is fully fine-tuned for 25 epochs with a batch size of 640 on 8 H100 GPUs, following our proposed recipe.
We set $N_{warm}=10k$ steps for our loss scheduling strategy as described in \cref{subsec:warm_start}. Other details can be found in the supplementary material.

\vspace{2pt}\noindent\textbf{Text-to-image generation details}. 
For text-to-image generation, we conduct quantitative and qualitative experiments based on Stable Diffusion 3.5 Medium~(SD3.5M)~\citep{esser2024scaling}, fine-tuning it to generate at a resolution of $1024\times1024$ with a base resolution of $512\times512$.
The SD3.5M uses a VAE with $f=8$ compression ratio and $C=16$ channels in the latent space.  Our structural latent adds an additional $D=16$ channels to the latent space, resulting in a total of 32 channels per token.
SD3.5 uses a MMDiT-X diffusion backbone with 2.5B parameters and patch size of $p=2$. 
We train \thevae for 10k steps with a batch size of 32 on SAM dataset~\citep{kirillov2023segment} and fine-tune the SD3.5M backbone for 20k steps with a batch size of 128 on a synthetic dataset generated from the base model using the prompts from DiffusionDB~\citep{wang2023diffusiondb}.
During fine-tuning, we use our proposed recipe and set $N_{warm}=5k$ steps. Due to compute constraints, we apply LoRA with a rank of 256 for all blocks in the DiT backbone, except for the patch embedder and the output layer.
For quantitative evaluation, we use the MJHQ-30K \cite{li2024playground} dataset and report CLIP-Score \cite{taited2023CLIPScore} and GenEval \cite{ghosh2023geneval} results; see supplementary material for other details.

\vspace{-4pt}
\subsection{ImageNet results}
\vspace{-2pt}

\input{tables/sd3}
\input{tables/alignment_loss_ablation}
\input{tables/ablation}

\noindent\textbf{Autoencoder evaluation}.
We first evaluate the reconstruction and generation performance of \thevae~compared to the base VA-VAE and other baselines.
For reconstruction, we report rFID~\cite{fid}, PSNR, LPIPS~\cite{zhang2018unreasonable} and SSIM~\cite{ssim} on the ImageNet validation set.
For generation, we train a DiT-XL for 20k steps \textbf{from scratch} under the same token budget and report FID-10k on ImageNet $512\times512$; no classifier-free guidance is applied for fair comparisons.
Our \thevae~achieves a better trade-off between reconstruction and generation compared to other autoencoder baselines (\cref{tab:imagenet-vae}), and its structured latent enables faster diffusion training despite having high channel dimensionality.

\vspace{2pt}\noindent\textbf{Diffusion fine-tuning evaluation}.
We compare class-to-image generation against state-of-the-art methods at $512\times512$ resolution.
Starting from VA-VAE with LightningDiT-XL ($p=1$), we compare against fine-tuning LightningDiT-XL with $p=2$ via the DC-Gen strategy~\cite{he2025dcgen}, and against training LightningDiT-XL with $p=2$ from scratch, annotated by an asterisk (*) in \cref{tab:imagenet}.
We also compare with other autoencoders for $512\times512$ generation that use the same or more tokens; our strategy shows better generation performance. Qualitative results in \cref{fig:res_imagenet} further demonstrate rich details and complex structures.

\vspace{-6pt}
\subsection{Text-to-image results}
\vspace{-3pt}

We evaluate text-to-image generation by applying our recipe to Stable Diffusion 3.5 Medium, using a base resolution of $512\times512$ to generate images at $1024\times1024$. We compare against state-of-the-art methods for $1024\times1024$ image generation.
As summarized in \cref{tab:sd3}, SD3.5 Medium fine-tuned with our \thevae~models achieves quantitative results comparable to its base model, while improving throughput by about $4\times$ at 1K resolution. For qualitative comparison, \cref{fig:compare_sd35} shows that our method produces images with more complex structures, richer details, and better prompt fidelity than the $512\times512$ SD3.5-M baseline.
We further evaluate scaling to $2048\times2048$. Naively running SD3.5-M at this resolution leads to clear quality degradation, with distorted large objects and occasional layout collapse. In contrast, our \thevae-augmented model generates high-quality $2048\times2048$ samples that better preserve both global structure and fine details, as shown in \cref{fig:compare_sd35_2k}.

\vspace{-6pt}
\subsection{Ablation studies}
\vspace{-3pt}

In this section, we conduct ablation studies to showcase different effects of the alignment loss weight, and validate our design choices by removing core components of our method one at a time. 
We conduct all ablation studies on the ImageNet for $512\times512$ class conditional generation based on the VA-VAE and LightningDiT-XL.
All VAE models are trained for 50 epochs and all DiT models are trained for 20 epochs for ablation studies. For fair comparisons, we do not apply any classifier-free guidance during sampling.

\vspace{2pt}\noindent\textbf{Impact of alignment loss weights}.
As shown in \cref{tab:alignment_loss_ablation}, we vary the alignment loss weight $\lambda_{align}$ in \cref{eq:total} and report reconstruction and generation performance.
Without alignment ($\lambda_{align}=0$), the model attains strong reconstruction fidelity but yields poor generation quality due to an unstructured latent space.
With a small weight ($\lambda_{align}=0.1$), generation quality improves substantially while reconstruction remains largely preserved.
With a large weight ($\lambda_{align}=1.0$), generation quality further improves, but reconstruction degrades due to over-regularization.
A moderate weight of $0.5$ achieves the best trade-off; we use $\lambda_{align}=0.5$ throughout.

\vspace{2pt}\noindent\textbf{Effectiveness of design choices}.
As shown in \cref{tab:ablation}, we ablate the three main components: (a)~alignment loss, (b)~zero-initialization, and (c)~loss scheduling.
Among the three, alignment and zero-init are crucial for effective generation, and loss scheduling yields a further improvement.

%% file: tables/imagenet.tex
\begin{table*}[t]
\centering
\footnotesize
\setlength{\tabcolsep}{6.5pt}
\begin{tabular}{l|c|cc|c|c|cc|c}
\toprule
\multirow{2}{*}{\textbf{Method}} &
\multirow{2}{*}{\makecell{\textbf{Training}\\\textbf{Regime}}} &
\multicolumn{2}{c|}{\textbf{Reconstruction}} &
\multirow{2}{*}{\makecell{\textbf{Token-nums}\\(\# of latent tokens)}} &
\multirow{2}{*}{\makecell{\textbf{Training}\\\textbf{Epoches}}} &
\multicolumn{2}{c|}{\textbf{FID-50k} $\downarrow$} &
\multirow{2}{*}{\makecell{\textbf{Inception}\\\textbf{Score} $\uparrow$}} \\
\cmidrule(lr){3-4} \cmidrule(lr){7-8}
& & \textbf{AutoEncoders} & \textbf{rFID} & & & \textbf{w/o CFG} & \textbf{w/ CFG} & \\
\midrule
DiT-XL$^{\dagger}$~\cite{peebles2023scalable}
  & Scratch
  & \multirow{2}{*}{SD-VAE (f8c4p2)} & \multirow{2}{*}{0.48} & \multirow{2}{*}{\texttt{32$\times$32}}
  & 2400 & 12.04 & 3.04 & 255.3 \\
REPA$^{\dagger}$~\cite{yu2024representation}
  & Scratch
  &  &  &  & 200 & - & 2.08 & 274.6 \\
\midrule
DiT-XL$^{\dagger}$~\cite{peebles2023scalable}
  & Scratch
  & \multirow{2}{*}{DC-AE (f32c32p1)} & \multirow{2}{*}{0.66} & \multirow{2}{*}{\texttt{16$\times$16}}
  & 2400 & 9.56 & 2.84 & 117.5 \\
DC-Gen-DiT-XL$^{\dagger}$~\cite{he2025dcgen}
  & Fine-tune
  &  &  &  & 80 & 8.21 & 2.22 & 122.5 \\
\midrule
LightningDiT-XL$^*$~\cite{yao2025vavae}
  & Scratch
  & \multirow{2}{*}{VA-VAE (f16c32p2)} & \multirow{2}{*}{0.50} & \multirow{2}{*}{\texttt{16$\times$16}}
  & 80 & 21.79 & 3.98 & 229.7 \\
LightningDiT-XL~\cite{yao2025vavae}
  & Fine-tune
  &  &  &  & 80 & 11.31 & 3.12 & 254.5 \\
\midrule
\multirow{2}{*}{\gape{\gape{\textbf{\thedit}}}} &
\multirow{2}{*}{\gape{\textbf{Fine-tune}}} &
\multirow{2}{*}{\gape{\textbf{\thevae(f32c128p1)}}} &
\multirow{2}{*}{\gape{\textbf{0.47}}} &
\multirow{2}{*}{\gape{\textbf{\texttt{16$\times$16}}}}
& {\textbf{\color{gray}25}} & {\textbf{\color{gray}6.04}} & {\textbf{\color{gray}2.07}} & {\textbf{\color{gray}277.6}} \\
& & & & & \textbf{80} & \textbf{4.84} & \textbf{1.68} & \textbf{314.3} \\
\bottomrule
\end{tabular}
\vspace{-9pt}
\caption{\textbf{ImageNet $512\times512$ comparison in training regime, efficiency, and performance.}
\textbf{Training Regime:} \emph{Scratch} trains the generator from random initialization for the target setting; \emph{Fine-tune} starts from a pretrained generator (or a closely-related pretrained checkpoint) and adapts it to the target setting (e.g., resolution/tokenizer/architecture change).
$^{\dagger}$ indicates numbers are directly copied from the corresponding papers; $^{*}$ follows the original paper's from-scratch setting.}
\label{tab:imagenet}
\vspace{-8pt}
\end{table*}

%% file: figures/res_imagenet.tex
\begin{figure*}
    \centering
    \includegraphics[width=1.0\linewidth]{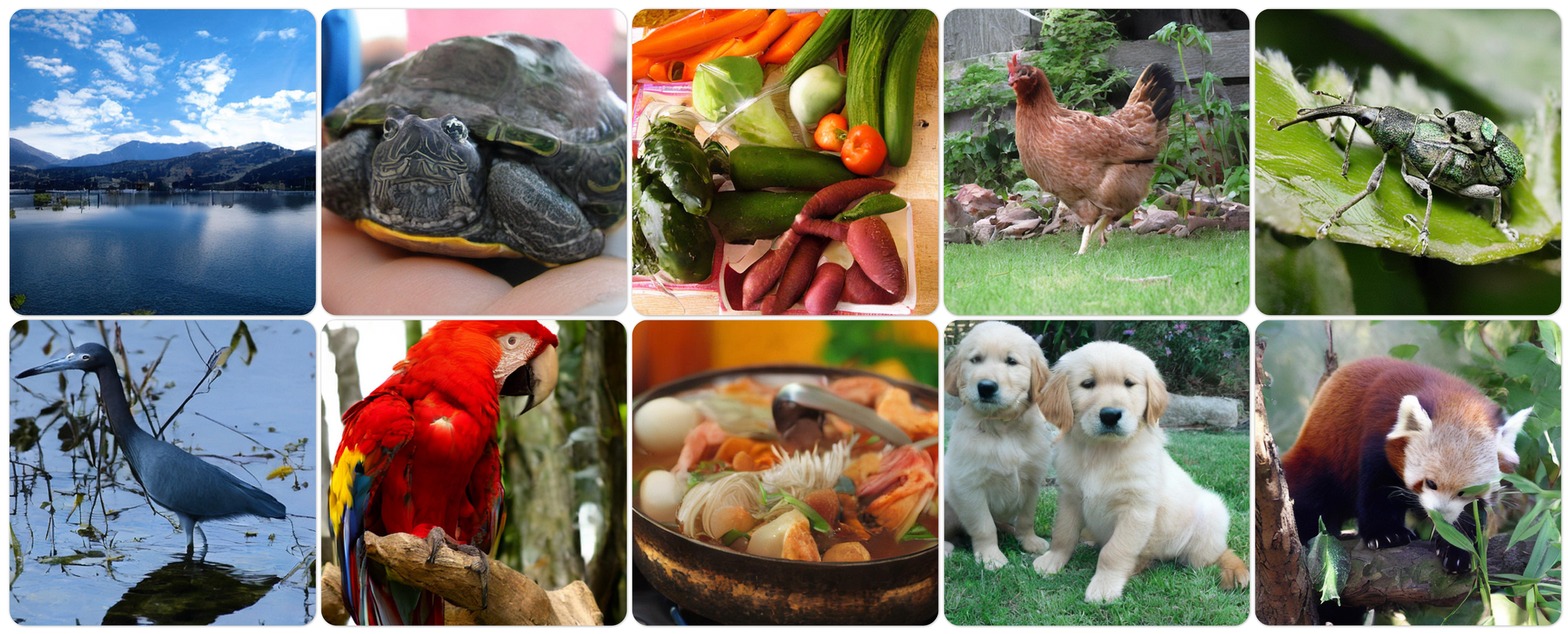}
    \vspace{-15pt}
    \caption{
    Qualitative samples from our model trained at $512\times512$ resolution on ImageNet.
    }
    \label{fig:res_imagenet}
    \vspace{-12pt}
\end{figure*}

%% file: tables/imagenet-vae.tex
\begin{table}[t]
\centering
\scriptsize
\setlength{\tabcolsep}{3.8pt}
\begin{tabular}{l|cccc|c}
\toprule
& \multicolumn{4}{c|}{\textbf{Reconstruction}} & \textbf{Generation} \\
\cmidrule(lr){2-5}\cmidrule(lr){6-6}
\textbf{Autoencoder} & \textbf{rFID} $\downarrow$ & \textbf{PSNR} $\uparrow$ & \textbf{LPIPS} $\downarrow$ & \textbf{SSIM} $\uparrow$ & \textbf{FID-10k} $\downarrow$ \\
\midrule
SD-VAE (f8c4p4)   & 0.48 & \textbf{29.22} & 0.13 & \textbf{0.79} & 58.17 \\
DC-AE (f32c32p1)  & 0.66 & 27.78 & 0.16 & 0.74 & 35.97 \\
VA-VAE (f16c32p2) & 0.50 & 28.43 & 0.13 & 0.78 & 44.65 \\
\rowcolor{blue!7}
\textbf{\thevae (f32c128p1)} & \textbf{0.47} & 28.53 & \textbf{0.12} & 0.78 & \textbf{31.51} \\
\bottomrule
\end{tabular}
\vspace{-9pt}
\caption{
    \textbf{Performance comparison of different autoencoders}. All generation models were trained from scratch.
}
\label{tab:imagenet-vae}
\vspace{-16pt}
\end{table}

%% file: figures/compare_sd35.tex
\begin{figure*}
    \centering
    \includegraphics[width=1.0\linewidth]{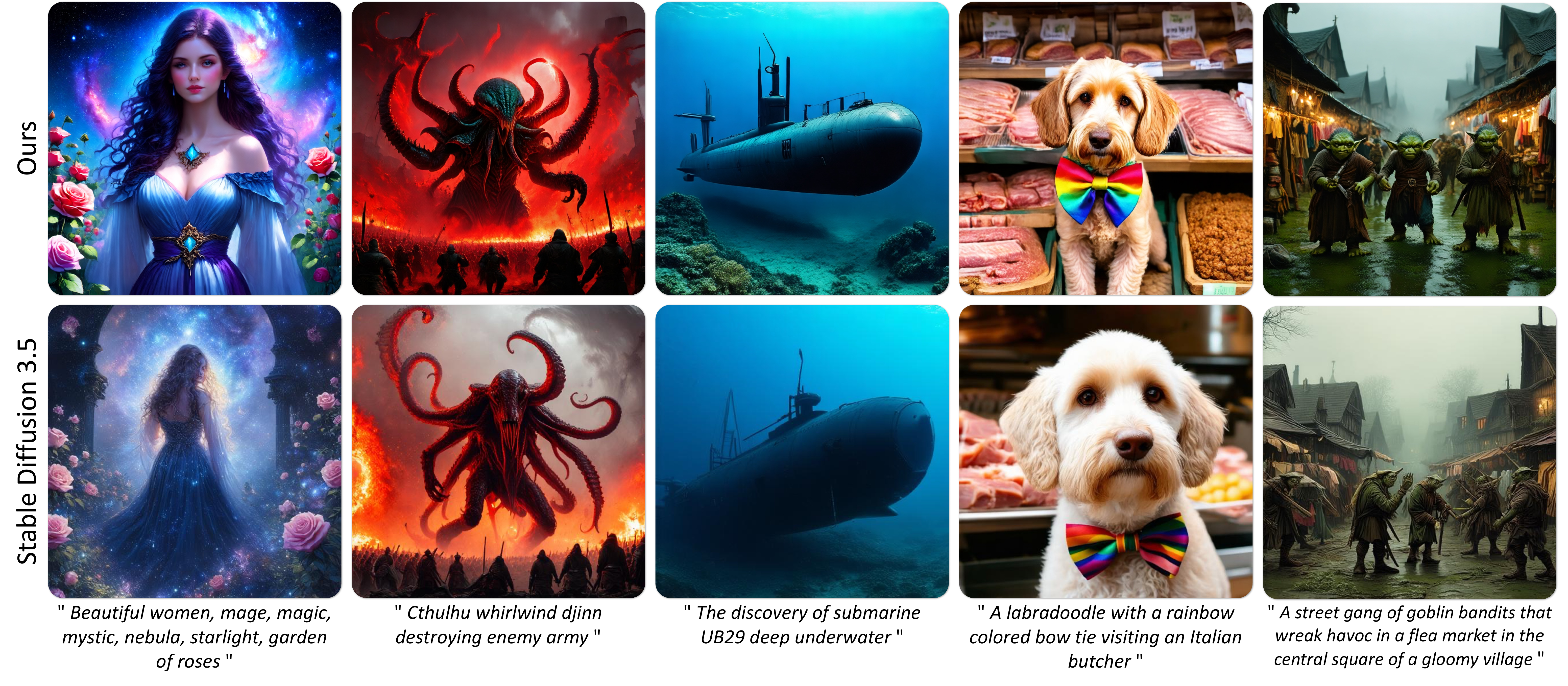}
    \vspace{-20pt}
    \caption{
    Comparison between our method ($1024\times1024$) and Stable Diffusion 3.5 ($1024\times1024$ by $512\times512$ upsample).
    }
    \label{fig:compare_sd35}
    \vspace{-8pt}
\end{figure*}

%% file: figures/compare_sd35_2k.tex
\begin{figure*}
    \centering
    \includegraphics[width=1.0\linewidth]{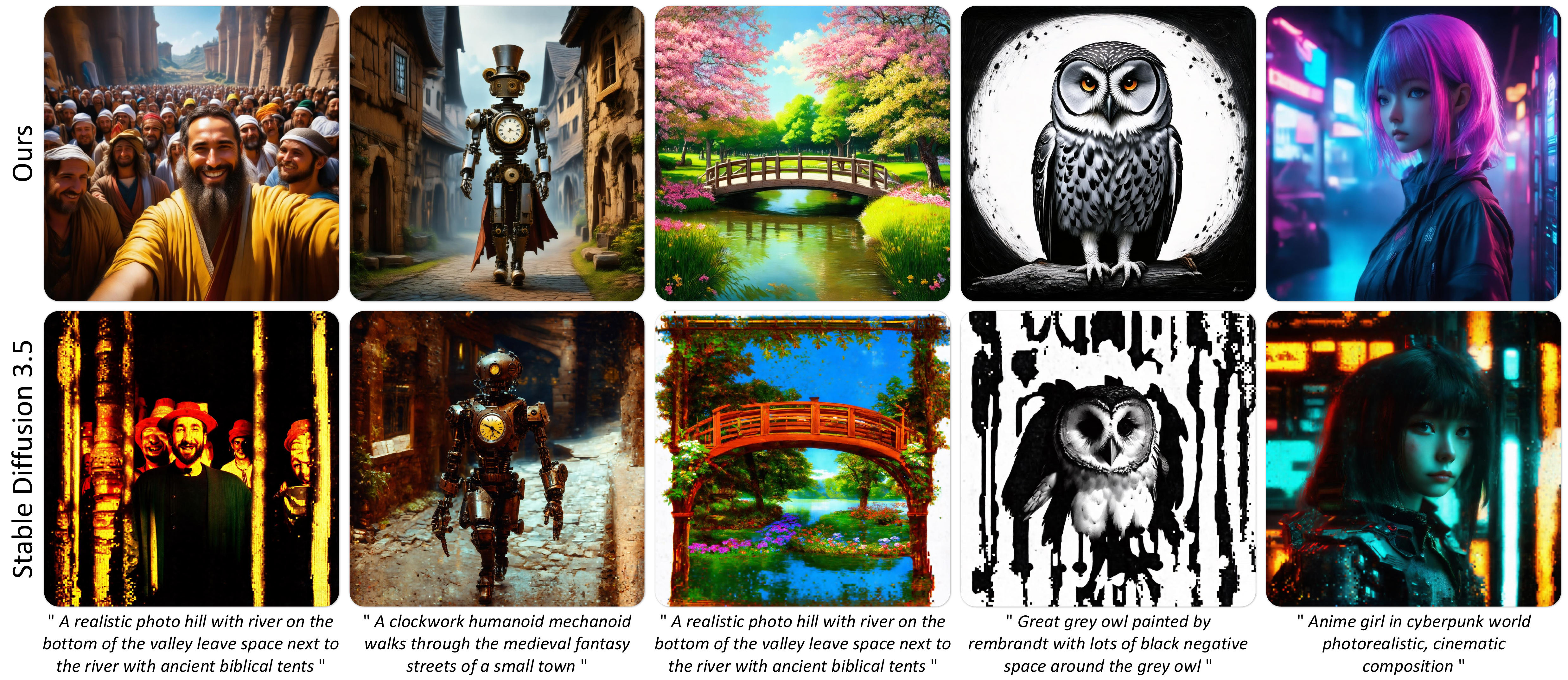}
    \vspace{-14pt}
    \caption{
    Comparison between our method and Stable Diffusion 3.5 on a $2048\times2048$ resolution. Zoom in for a better view. }
    \label{fig:compare_sd35_2k}
    \vspace{-16pt}
\end{figure*}

%% file: tables/sd3.tex
\begin{table*}[t]
\centering
\footnotesize
\setlength{\tabcolsep}{7.6pt}
\begin{tabular}{l|c|c|c|c|ccc}
\toprule
\textbf{Method} &
\textbf{AutoEncoders} &
\makecell{\textbf{Token-nums}\\(\# of latent tokens)} &
\makecell{\textbf{Params}\\\textbf{(B)}} &
\makecell{\textbf{Throughput}\\(img/s)} &
\textbf{FID $\downarrow$} &
\makecell{\textbf{CLIP}\\\textbf{Score} $\uparrow$} &
\textbf{GenEval $\uparrow$} \\
\midrule
PixArt-$\Sigma$        & —                   & \texttt{64$\times$64} & 0.6 & 0.40 & 6.15 & 28.26 & 0.54 \\
Hunyuan-DiT            & —                   & \texttt{64$\times$64} & 1.5 & 0.05 & 6.54 & 28.19 & 0.63 \\
SANA-1.5               & DC-AE(f32c32p1)      & \texttt{32$\times$32} & 4.8 & 0.26 & 5.99 & 29.23 & 0.80 \\
FLUX-dev               & FLUX-VAE(f8c16p2)    & \texttt{64$\times$64} & 12  & 0.04 & 10.15 & 27.47 & 0.67 \\
SD3-medium             & SD3-VAE (f8c16p2)    & \texttt{64$\times$64} & 2.0 & 0.36 & 11.92 & 27.83 & 0.62 \\
\midrule
SD3.5-medium           & SD3-VAE (f8c16p2)    & \texttt{64$\times$64} & 2.5 & 0.25 & 10.31 & 29.74 & 0.63 \\
SD3.5-medium$^{\dagger}$ & SD3-VAE (f8c16p2)  & \texttt{32$\times$32} & 2.5 & 1.03 & 12.04 & 30.17 & 0.63 \\
\rowcolor{blue!7}
\textbf{Ours (SD3.5-M + \thevae)} &
\textbf{\thevae~(f16c32p2)}       & \textbf{\texttt{32$\times$32}} & 2.5 & 1.03 & 10.91 & 31.91 & 0.64 \\
\bottomrule
\end{tabular}
\vspace{-9pt}
\caption{\textbf{Comparison of our method with SOTA approaches in efficiency and performance.}
FID and CLIP Score are reported on MJHQ-30K (1024$\times$1024).
Throughput is measured on a single A100 GPU (BF16, batch size 10).
\textbf{Data sources:} the first five baselines (PixArt-$\Sigma$, Hunyuan-DiT, SANA-1.5, FLUX-dev, and SD3-medium) are copied from \cite{xie2025sana} under the same evaluation protocol.}
\label{tab:sd3}
\vspace{-10pt}
\end{table*}

%% file: tables/alignment_loss_ablation.tex
\begin{table}[t]
\centering
\footnotesize
\setlength{\tabcolsep}{2.9pt}
\begin{tabular}{c|cccc|c}
\toprule
\textbf{Alignment } & \multicolumn{4}{c|}{\textbf{Reconstruction}} & \textbf{Generation} \\
\cmidrule(lr){2-5}\cmidrule(lr){6-6}
\textbf{loss weight} $\boldsymbol{\lambda_{\text{align}}}$ & \textbf{rFID} $\downarrow$ & \textbf{PSNR} $\uparrow$ & \textbf{LPIPS} $\downarrow$ & \textbf{SSIM} $\uparrow$ & \textbf{FID-10k} $\downarrow$ \\
\midrule
0.0 & 0.59 & 29.23 & 0.11 & 0.80 & 16.37 \\
0.1 & 0.55 & 28.70 & 0.12 & 0.79 & 9.58 \\
\rowcolor{blue!7} 
0.5 & 0.47 & 28.53 & 0.12 & 0.78 & 9.27 \\
1.0 & 0.63 & 27.90 & 0.14 & 0.76 & \textbf{9.23} \\
\bottomrule
\end{tabular}
\vspace{-8pt}
\caption{\textbf{Ablation on alignment-loss weight.} Increasing $\lambda_{\text{align}}$ slightly degrades reconstruction (higher rFID / LPIPS, lower PSNR / SSIM) but improves generation quality (lower gFID), with the best trade-off at a moderate weight.}
\label{tab:alignment_loss_ablation}
\vspace{-8pt}
\end{table}

%% file: tables/ablation.tex
\begin{table}[t]
\centering
\scriptsize
\setlength{\tabcolsep}{3pt}
\begin{tabular}{lccc|c}
\toprule
\textbf{Method} & \textbf{Alignment} & \textbf{Zero Init} & \textbf{Weight Scheduler} & \textbf{FID-10k $\downarrow$} \\
\midrule
\textbf{Ours (full)}              & \cmark & \cmark & \cmark & \textbf{9.27} \\
\midrule
w/o alignment                & \xmark & \cmark & \cmark & 16.37 \\
w/o zero init                     & \cmark & \xmark & \cmark & 29.73 \\
w/o weight scheduler      & \cmark & \cmark & \xmark & 9.80 \\
\bottomrule
\end{tabular}
\vspace{-9pt}
\caption{\textbf{Ablation on three components}. Our full model enables all three (\cmark); each ablation disables exactly one component (\xmark).}
\label{tab:ablation}
\vspace{-15pt}
\end{table}

%% file: sec/5_final.tex
\vspace{-6pt}
\section{Conclusion and Limitations}
\vspace{-4pt}

Our method provides a simple and efficient recipe to increase the effective compression ratio of a pretrained VAE while keeping the token count unchanged, enabled by detail alignment. This approach does not require expensive retraining.
We showcase promising results for generic text-to-image tasks, where our method enables higher-resolution generation with the same number of visual tokens.
However, our work has several limitations. First, we deliberately choose our current detail alignment loss due to its simplicity; there may be better alternatives. Second, given our compute budget, we have not yet evaluated full fine-tuning on SD3.5 or applied our method to more recent but costly backbones such as FLUX. Finally, as a proof-of-concept, our method currently uses synthetic data for fine-tuning. Therefore, our generated images are less photorealistic than SD3.5's native generation at $1024\times 1024$. We leave these directions for future work.

%% file: sec/X_suppl.tex
\clearpage
\setcounter{page}{1}
\onecolumn
\begin{center}
    {\Large \textbf{DA-VAE: Plug-in Latent Compression for Diffusion via Detail Alignment --- Supplementary Material}}
\end{center}

\label{sec:rationale}
\noindent
This supplementary material provides implementation details and additional analyses.
In particular, it
\begin{itemize}
    \item \cref{sec:training-details} summarizes the training and
    sampling hyperparameters used in all experiments;
    \item \cref{sec:arch-details} describes how to instantiate
    DA-VAE on top of a pretrained VAE tokenizer, using SD3-VAE as a
    concrete example;
    \item \cref{sec:decoder-sensitivity} and \cref{sec:training-dynamics}
    verify that the decoder and the diffusion backbone actually make use
    of the extra detail latent channels, rather than ignoring them;
    \item \cref{sec:sr-comparison} provides a detailed comparison of
    DA-VAE against super-resolution post-processing baselines;
    \item \cref{sec:spectral-analysis} presents a frequency-domain
    analysis of the base and detail latents;
    \item \cref{sec:additional-results} presents additional qualitative
    results for DA-VAE enhanced Stable Diffusion 3.5 Medium.
\end{itemize}

\section{Training and sampling hyperparameters}
\label{sec:training-details}

\cref{tab:hyperparameter_davae_dit} lists the optimization and sampling configurations used in all our experiments.

For ImageNet class-to-image experiments with LightningDiT-XL, we largely follow the training recipe of Yao et al.~\cite{yao2025vavae}, adjusting only the learning rate, batch size, and loss weights to accommodate our higher-compression DA-VAE latent space.
DA-VAE is trained with AdamW and a relatively small KL weight, while $\lambda_{\text{align}}$ is set to a moderate value to balance reconstruction and generation quality.
For SD3.5-M, we use a smaller batch size and slightly different loss weights $(\lambda_L, \lambda_1, \lambda_{\text{adv}}, \lambda_{\text{KL}}, \lambda_{\text{align}})$ to stabilize high-resolution reconstruction.
During DiT fine-tuning, the gradual loss scheduling down-weights the detail-latent loss for the first $N_{\text{warm}}$ steps (10k for LightningDiT-XL; 5k for SD3.5-M), after which it is ramped up to full weight.
We maintain an EMA of the DiT parameters with decay $0.999$ throughout.
For sampling, we use 250 diffusion steps with CFG scale $4.0$ on ImageNet and 30 steps with guidance scale $2.5$ for SD3.5-M.

\input{tables/hyper_params}

\section{DA-VAE architecture}
\label{sec:arch-details}

\cref{fig:davae} illustrates how we turn a SD3-VAE into DA-VAE.
We \emph{borrow} the overall encoder and decoder backbone architectures from SD3-VAE, but remove their original feature-to-latent and latent-to-feature heads and replace them with our own downsampling and upsampling blocks.
The resulting encoder $E_f$ and decoder $D_f$ are therefore retrained as part of DA-VAE.

Concretely, the SD3-VAE encoder produces an intermediate feature map $F \in \mathbb{R}^{512 \times H \times W}$.
In the original SD3-VAE, a shallow head directly maps $F$ to a latent $z_{\text{sd3}} \in \mathbb{R}^{16 \times H \times W}$.
In our design, we discard this head and instead attach a small downsampling module that further reduces the spatial resolution of $F$ while keeping the channel dimension fixed (e.g., a stack of strided $3 \times 3$ conv blocks).
This yields a more compressed detail latent $z_d \in \mathbb{R}^{16 \times (H/s) \times (W/s)}$.
We then concatenate $z_d$ with the base latent $z$ (the original SD3-VAE feature of the downsampled base image) to form the structured latent $(z, z_d)$ used by our DiT.

The decoder side is modified symmetrically.
Instead of feeding the original SD3-VAE latent $z_{\text{sd3}}$ into a latent-to-feature stem, we concatenate our base and detail latents along the channel dimension and apply a lightweight upsampling block (e.g., pixel shuffle) that inverts the encoder's spatial downsampling.
A $3 \times 3$ convolution then maps the upsampled latent back to a $512 \times H \times W$ feature map, which is passed through the SD3-VAE decoder backbone $D_f$ for reconstruction.

In summary, DA-VAE keeps the deep convolutional backbone structure of SD3-VAE but replaces its shallow latent heads with our own downsampling/upsampling design, enabling a higher-compression latent space with an explicit separation between base and detail channels.
All components, including the reused backbone blocks, are trained end-to-end under our DA-VAE objective.

\input{figures/downsample_block}

\section{Decoder sensitivity to the detail latent}
\label{sec:decoder-sensitivity}

\input{tables/vae-ablation}

We evaluate the sensitivity of the decoder to the detail latent on the ImageNet validation set.
Starting from a trained \thevae, we fix the base latent $z$ and modify the detail latent $z_d$ in two ways:
(i) we replace $z_d$ with i.i.d.\ Gaussian noise $\mathcal{N}(0,I)$ (\emph{Base + random detail}); and
(ii) we set $z_d$ to zero (\emph{Base + zero detail}).
The quantitative reconstruction metrics are summarized in \cref{fig:decoder-detail-ablation-metric}, and representative reconstructions are visualized in \cref{fig:decoder-detail-ablation-qual}.

Randomizing $z_d$ leads to clearly invalid reconstructions with high rFID, low PSNR, and severe artifacts such as distorted faces and unreadable text, indicating that the decoder cannot simply ignore the detail channels.
Zeroing $z_d$ produces structurally plausible but over-smoothed images: edges become soft and fine textures disappear.
In contrast, the full model using both $z$ and $z_d$ recovers both global structure and high-frequency details.

These observations confirm that the learned detail latent encodes semantically meaningful fine-grained information.
Consequently, during fine-tuning the DiT must also learn to generate $z_d$ correctly; otherwise the final high-resolution samples would lack sharp details even if the base latent is well modeled.

\section{Training dynamics of SD3.5-M fine-tuning}
\label{sec:training-dynamics}

\input{figures/train_loss}

\noindent\cref{fig:loss} visualizes the optimization behaviour when fine-tuning SD3.5-M from $512\times512$ to $1024\times1024$ resolution with our DA-VAE.
We plot the \emph{unweighted} diffusion loss on the base latent and on the detail latent, i.e., the true per-token MSE before applying the scheduling weight $w(n)$ described in the main paper.
For each branch we show both the raw loss and its exponential moving average (EMA).

Two trends are worth noting.
First, the base-latent loss stays relatively low throughout training, while the detail-latent loss starts much higher and gradually decreases—fine-tuning primarily teaches the model to predict the new detail channels, leveraging the well-trained prior in the base latent.
Second, comparing the two plots shows the effect of latent alignment: without alignment the detail-latent loss plateaus at a high value, whereas with alignment it decreases steadily and eventually falls \emph{below} the base-latent loss, confirming that aligned latents form a more learnable distribution that the DiT can effectively exploit.

\section{Comparison with Super-Resolution Post-Processing}
\label{sec:sr-comparison}

A natural question is whether one could achieve similar results by first generating a low-resolution image and then applying a learned super-resolution (SR) model.
We argue that DA-VAE is superior in two key aspects.

\noindent\textbf{Joint modeling vs.\ conditional upsampling.}
A two-stage SR pipeline factorizes the high-resolution distribution as $P(x_{\text{high}}) \approx P(x_{\text{low}})\,P(x_{\text{high}} \mid x_{\text{low}})$.
Once the $512$px model has sampled $x_{\text{low}}$, the global composition (e.g., layout, object counts) is largely fixed; the SR model can only refine local appearance and cannot reliably correct missing objects or compositional errors.
In contrast, DA-VAE models the joint distribution $P(x_{\text{high}})$ natively, yielding better structural fidelity and text alignment, as reflected by the higher GenEval-\textit{Count} and CLIP-Score in \cref{tab:sr-comparison}.

\noindent\textbf{Inference latency.}
SR requires a cascaded second-stage inference pass, adding non-trivial latency (e.g., SeedVR2 roughly doubles total inference time compared to the $512$px baseline).
DA-VAE generates high-resolution images in a single forward pass, matching the throughput of the $512$px baseline.

\noindent\cref{tab:sr-comparison} summarizes quantitative results and \cref{fig:sr-comparison} shows qualitative examples.
In the counting example, $512$px generation produces an incorrect count that SR methods cannot fix, whereas DA-VAE generates the correct number of objects directly.
In the scene example, SR sharpens local textures but preserves a simplified layout, while DA-VAE produces richer global structure.

\begin{center}
\begin{minipage}{0.7\linewidth}
\setlength{\tabcolsep}{5pt}
\captionsetup{type=table}
\caption{Comparison of DA-VAE with super-resolution post-processing baselines. All methods use the same 512$\times$512 SD3.5-M backbone. Throughput in img/s on a single H100.}
\label{tab:sr-comparison}
\small
\centering
\begin{tabular}{lcccc}
\toprule
Method & FID$\downarrow$ & GenEval-\textit{Count}$\uparrow$ & CLIP-Score$\uparrow$ & Throughput$\uparrow$ \\
\midrule
512 + Bilinear         & 12.04          & 0.55          & 30.17          & 1.03 \\
512 + SeedVR2          & \textbf{10.48} & 0.55          & 30.19          & 0.45 \\
512 + FMBoost          & 11.02          & 0.55          & 30.16          & 0.52 \\
\textbf{DA-VAE (Ours)} & 10.91          & \textbf{0.60} & \textbf{31.91} & \textbf{1.03} \\
\bottomrule
\end{tabular}
\end{minipage}
\end{center}

\begin{center}
\begin{minipage}{0.7\linewidth}
\captionsetup{type=figure}
\includegraphics[width=\linewidth]{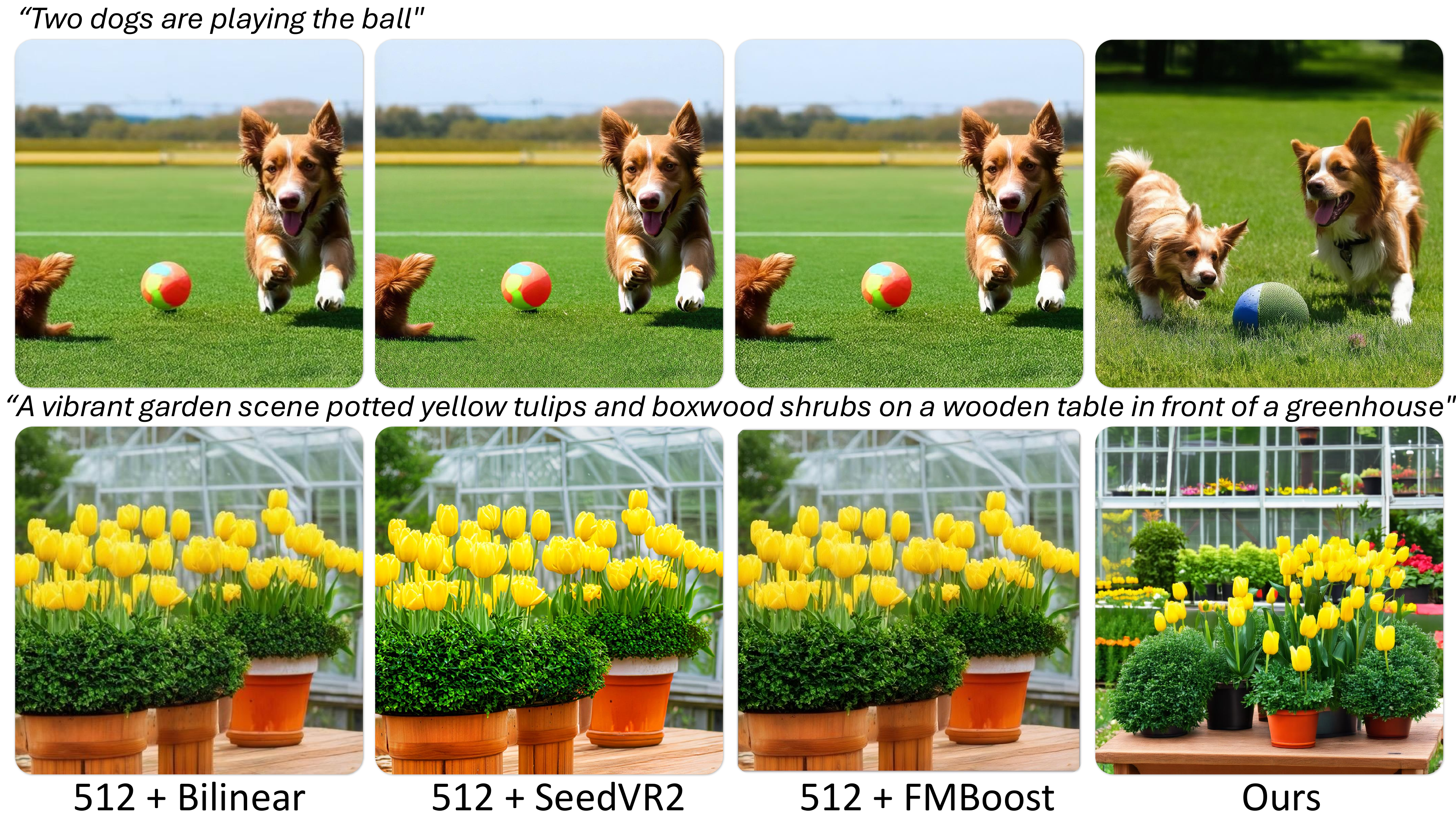}
\caption{Qualitative comparison of DA-VAE vs.\ SR baselines.
  \textit{Top:} a counting prompt where 512px generation gets the count wrong and SR cannot fix it.
  \textit{Bottom:} a scene prompt where SR only sharpens local textures while DA-VAE produces richer global structure.}
\label{fig:sr-comparison}
\end{minipage}
\end{center}

\section{Frequency-Domain Analysis of Base and Detail Latents}
\label{sec:spectral-analysis}

To verify that the detail latent $\mathbf{z}_d$ encodes genuinely complementary high-frequency information—rather than simply duplicating the base latent $\mathbf{z}$—we compute the radial power spectrum of each latent channel and average across channels and images from the ImageNet validation set.

\cref{fig:spectral} plots the resulting spectral energy as a function of spatial frequency.
The base latent $\mathbf{z}$ concentrates energy at low frequencies, consistent with its role in capturing global structure, while $\mathbf{z}_d$ exhibits substantially higher energy in the mid-to-high frequency bands, confirming that it captures fine textures and edges absent from $\mathbf{z}$.
This is consistent with \cref{sec:decoder-sensitivity}: zeroing $\mathbf{z}_d$ produces over-smoothed reconstructions precisely because this high-frequency content is lost.
Despite this complementarity, the alignment loss prevents $\mathbf{z}_d$ from collapsing into a trivial copy of $\mathbf{z}$: the two latents differ in both spectral content and spatial statistics, making them jointly necessary for full-resolution reconstruction.

\begin{center}
\begin{minipage}{0.7\linewidth}
\captionsetup{type=figure}
\includegraphics[width=\linewidth]{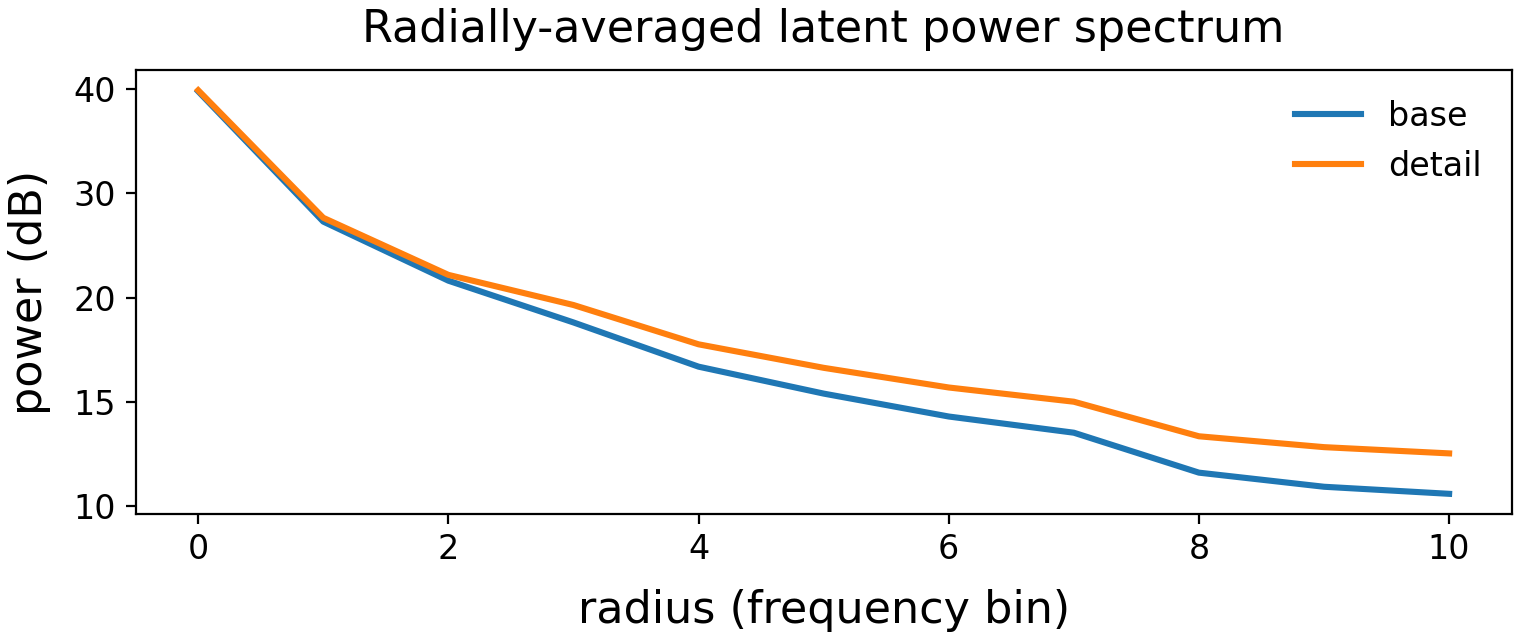}
\caption{Radial power spectrum of the base latent $\mathbf{z}$ and detail latent $\mathbf{z}_d$ averaged over ImageNet validation images.
The detail latent carries substantially more high-frequency energy, confirming that it encodes complementary fine-grained information rather than duplicating the base.}
\label{fig:spectral}
\end{minipage}
\end{center}

\section{Additional qualitative results}
\label{sec:additional-results}

To further demonstrate the effectiveness of our method, \cref{fig:additional_results} presents additional qualitative results of DA-VAE–enhanced Stable Diffusion 3.5 Medium (SD3.5-M) for text-to-image generation.
To improve realism, we further fine-tune SD3.5-M with our model for 5K steps on 500K images generated by Flux~\cite{flux} using prompts collected by~\cite{chen2025pi}.

\input{figures/visual}

%% file: tables/hyper_params.tex
\begin{table}[h]
\vspace{-6pt}
\centering
\resizebox{1.0\textwidth}{!}{
\begin{tabular}{c|c|c|c}
\toprule
\textbf{Stage} & \textbf{Hyper-parameter} & \textbf{lightningDiT-XL \cite{yao2025vavae}, Class-to-image} & \textbf{SD3.5-M \cite{sd35}, Text-to-image} \\
\midrule
\multirow{6}{*}{DA-VAE Training}
    & learning rate   & 1e-4 & 1e-4 \\
    & batch size      & 128 & 16 \\
    & training steps  & 100K & 10K \\
    & optimizer       & AdamW, betas=[0.5, 0.9] & AdamW, betas=[0.9, 0.999] \\
    & loss weights $(\lambda_L,\lambda_1,\lambda_{\mathrm{adv}},\lambda_{KL},\lambda_{\mathrm{align}})$
                      & (1.0, 1.0, 0.1, 1e-6, 0.5) & (1.0, 2.0, 0.1, 1e-7, 1.0) \\
\midrule
\multirow{7}{*}{DiT Fine-Tuning}
    & learning rate   & 2e-4 & 1e-4 \\
    & Gradual loss scheduling steps    & 10K & 5K \\
    & batch size      & 640 & 128 \\
    & training steps  & 140K & 10K \\
    & optimizer       & AdamW, betas=[0.9, 0.95] & AdamW, betas=[0.9, 0.999] \\
    & EMA decay       & 0.999 & 0.999 \\
\midrule
\multirow{4}{*}{Sampling for Generation}
    & \# sampling steps    & 250 & 30 \\
    & CFG / guidance scale & 4.0 & 2.5 \\
    & CFG interval start           & 0.2 & - \\
    & timestep shift           & 0.3 & - \\
\bottomrule
\end{tabular}
}
\vspace{-4pt}
\caption{\textbf{Training and sampling hyperparameters for lightningDiT-XL and SD3.5-M.}}
\label{tab:hyperparameter_davae_dit}
\vspace{-4pt}
\end{table}

%% file: figures/downsample_block.tex
\begin{figure*}[h]
    \vspace{-10pt}
    \centering
    \includegraphics[width=\linewidth]{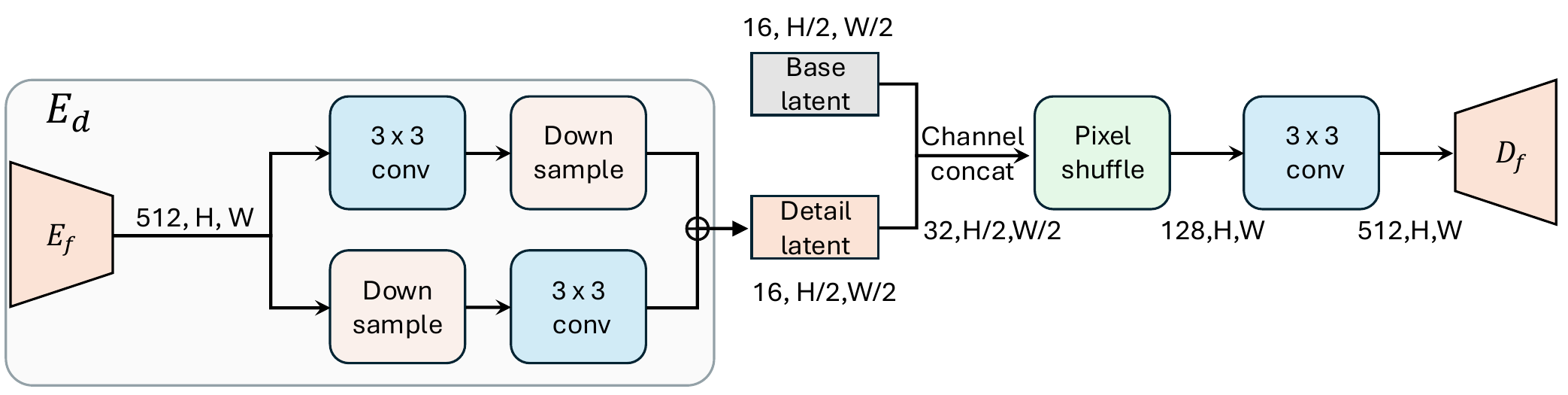}
    \caption{
   \textbf{DA-VAE architecture instantiated on SD3-VAE.
    }We reuse the convolutional encoder $E_f$ and decoder $D_f$ blocks from
SD3-VAE, but remove its original feature-to-latent and latent-to-feature
heads. Instead, a lightweight downsampling module maps the shared
$512\times H\times W$ feature map to a more compressed detail latent $z_d$
and a parallel base latent $z$ of the same shape, while a symmetric
upsampling module concatenates $(z, z_d)$, upsamples them back to
$512\times H\times W$, and feeds the result into the reused decoder
backbone. This yields a higher-compression latent space with explicit
base and detail channels, while keeping most of the VAE architecture
intact.}
    \label{fig:davae}
\end{figure*}

%% file: tables/vae-ablation.tex
\begin{figure}[h]
\centering
\footnotesize
\setlength{\extrarowheight}{5pt}
\setlength{\tabcolsep}{5pt}

\begin{minipage}[t]{0.55\linewidth}
\centering
\vspace{0pt}
\resizebox{\linewidth}{!}{%
\begin{tabular}{l|cccc}
\toprule
& \multicolumn{4}{c}{\textbf{Reconstruction (ImageNet val)}} \\
\cmidrule(lr){2-5}
\textbf{Decoder variant} & \textbf{rFID} $\downarrow$ & \textbf{PSNR} $\uparrow$ & \textbf{LPIPS} $\downarrow$ & \textbf{SSIM} $\uparrow$ \\
\midrule
\rowcolor{blue!7}
\textbf{Full (base + detail)} & \textbf{0.47} & \textbf{28.53} & \textbf{0.12} & \textbf{0.78} \\
Base + random detail          & 8.25        & 23.67          & 0.30          & 0.62          \\
Base + zero detail            & 2.93          & 24.71          & 0.25          & 0.63          \\
\bottomrule
\end{tabular}
}
\subcaption{Reconstruction metrics on the ImageNet validation set.}
\label{fig:decoder-detail-ablation-metric}
\end{minipage}
\hfill
\begin{minipage}[t]{0.43\linewidth}
\centering
\vspace{0pt}
\includegraphics[width=\linewidth]{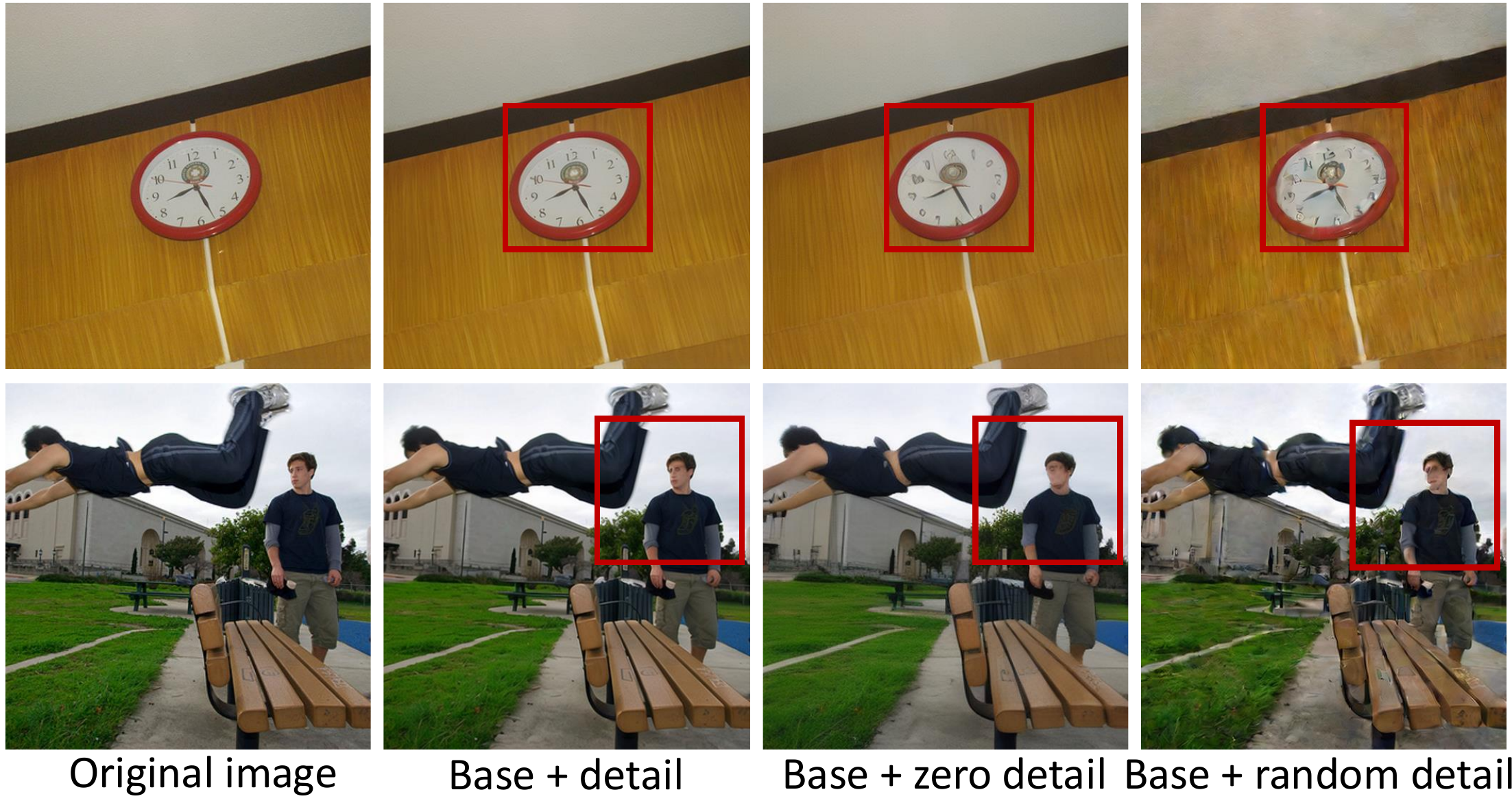}
\subcaption{Example reconstructions on ImageNet. Best for zoom-in view.}
\label{fig:decoder-detail-ablation-qual}
\end{minipage}

\caption{\textbf{Ablation on detail channels in the \thevae\ decoder on
ImageNet.} (a) Reconstruction metrics for different decoder variants.
(b) Visual examples showing that randomizing or zeroing the detail latent
either destroys the image or removes fine-grained details such as faces
and text. Please zoom in for best view.}
\label{fig:decoder-detail-ablation}
\end{figure}

%% file: figures/train_loss.tex
\begin{center}
\begin{minipage}{0.9\linewidth}
\captionsetup{type=figure}
\includegraphics[width=\linewidth]{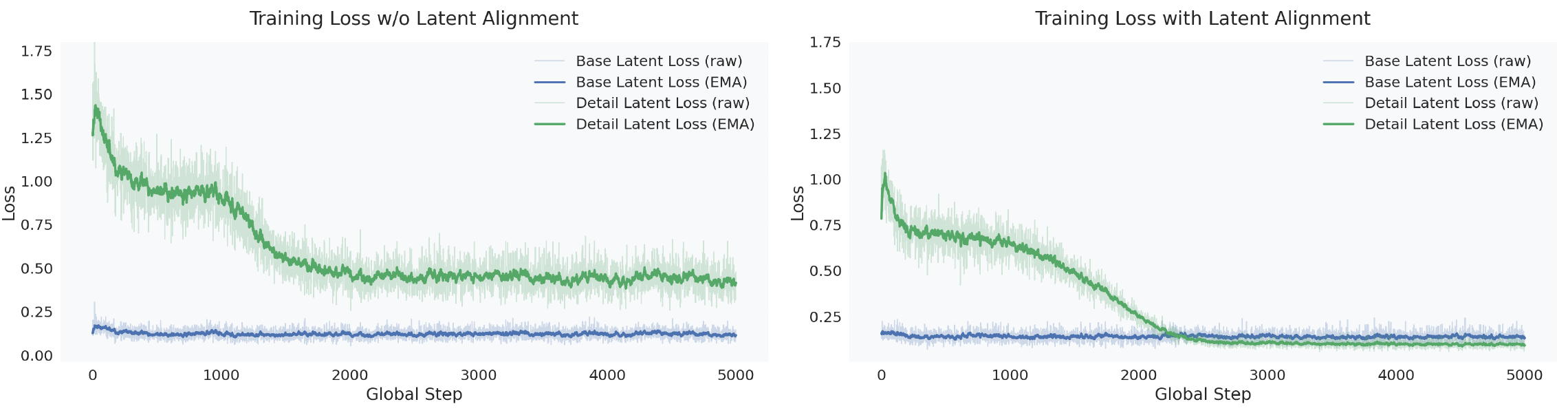}
\caption{\textbf{Training loss curves for SD3.5-M fine-tuning with and without latent alignment.}
We plot the unweighted diffusion loss on the base latent (blue) and the detail latent (green), showing both the raw loss (faint) and its EMA (solid).
\emph{Left:} without alignment, the detail-latent loss decreases slowly and stays significantly higher than the base-latent loss.
\emph{Right:} with alignment, optimization is more stable and the detail-latent loss eventually falls below the base-latent loss, indicating that the DiT has learned a well-structured distribution over the extra detail channels.}
\label{fig:loss}
\end{minipage}
\end{center}

%% file: figures/visual.tex
\begin{figure*}[t]
    \centering
    \includegraphics[width=\linewidth]{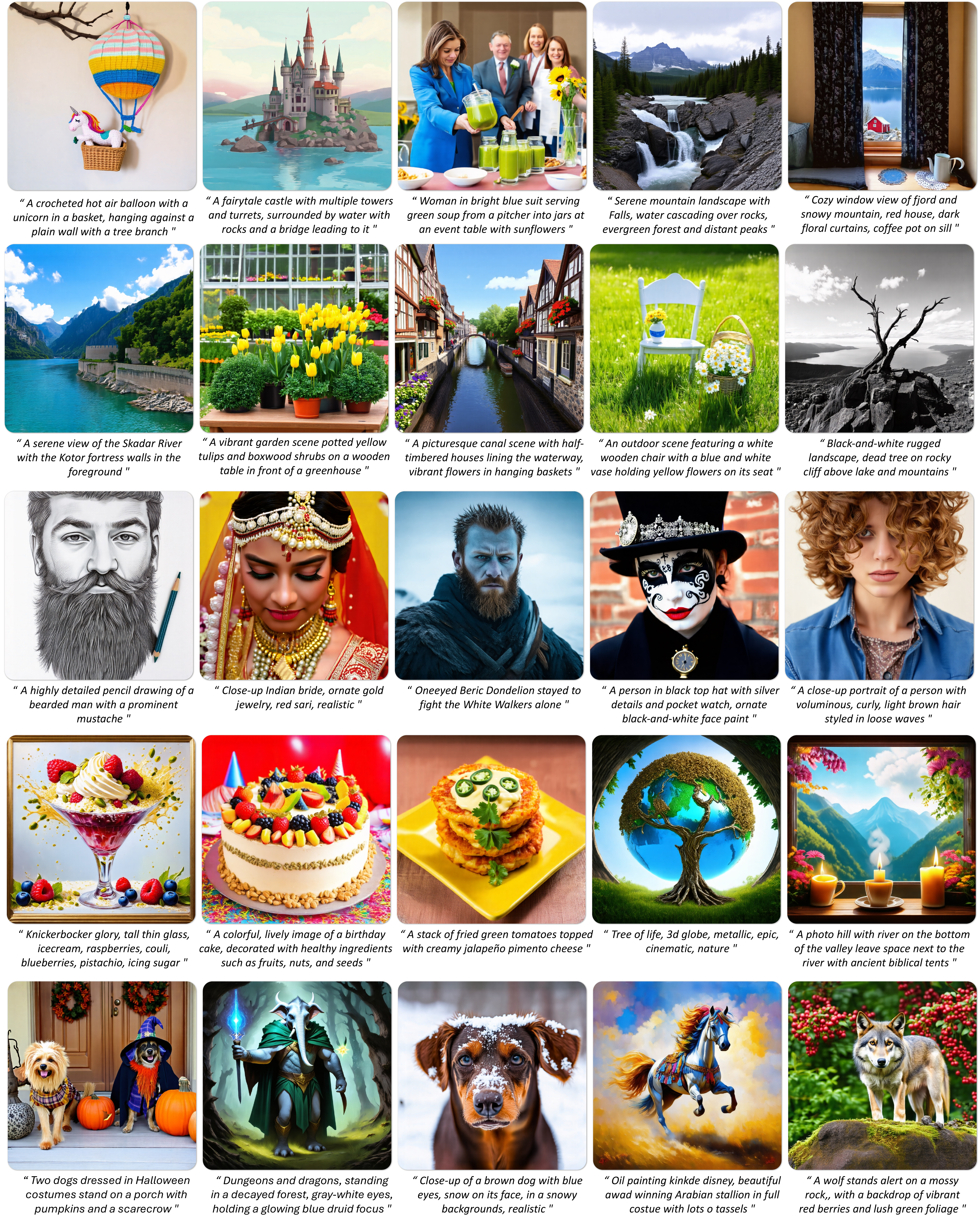}
    \caption{
   \textbf{Genearated examples by our DA-VAE enhanced SD3.5-M.
    } Please zoom in for best view.}
    \label{fig:additional_results}
\end{figure*}

%% file: main.bbl
\begin{thebibliography}{63}
\providecommand{\natexlab}[1]{#1}
\providecommand{\url}[1]{\texttt{#1}}
\expandafter\ifx\csname urlstyle\endcsname\relax
  \providecommand{\doi}[1]{doi: #1}\else
  \providecommand{\doi}{doi: \begingroup \urlstyle{rm}\Url}\fi

\bibitem[Ansel et~al.(2024)Ansel, Yang, He, Gimelshein, Jain, Voznesensky, Bao,
  Bell, Berard, Burovski, et~al.]{ansel2024pytorch}
Jason Ansel, Edward Yang, Horace He, Natalia Gimelshein, Animesh Jain, Michael
  Voznesensky, Bin Bao, Peter Bell, David Berard, Evgeni Burovski, et~al.
\newblock Pytorch 2: Faster machine learning through dynamic python bytecode
  transformation and graph compilation.
\newblock In \emph{Proceedings of the 29th ACM International Conference on
  Architectural Support for Programming Languages and Operating Systems, Volume
  2}, 2024.

\bibitem[Bao et~al.(2023)Bao, Nie, Xue, Cao, Li, Su, and Zhu]{bao2023all}
Fan Bao, Shen Nie, Kaiwen Xue, Yue Cao, Chongxuan Li, Hang Su, and Jun Zhu.
\newblock All are worth words: A {ViT} backbone for diffusion models.
\newblock In \emph{CVPR}, 2023.

\bibitem[{Black Forest Labs}(2024)]{flux}
{Black Forest Labs}.
\newblock Flux.
\newblock \url{https://github.com/black-forest-labs/flux}, 2024.
\newblock Accessed 2025-11-20.

\bibitem[Chen et~al.(2025{\natexlab{a}})Chen, Bi, Tan, Zhang, Zhang, Li, Xiong,
  Zhang, and Zhang]{chen2025aligning}
Bowei Chen, Sai Bi, Hao Tan, He Zhang, Tianyuan Zhang, Zhengqi Li, Yuanjun
  Xiong, Jianming Zhang, and Kai Zhang.
\newblock Aligning visual foundation encoders to tokenizers for diffusion
  models.
\newblock \emph{arXiv preprint arXiv:2509.25162}, 2025{\natexlab{a}}.

\bibitem[Chen et~al.(2024{\natexlab{a}})Chen, Wang, Li, Sun, Chen, Liu, Wang,
  Raj, Liu, and Barsoum]{chen2024softvq}
Hao Chen, Ze Wang, Xiang Li, Ximeng Sun, Fangyi Chen, Jiang Liu, Jindong Wang,
  Bhiksha Raj, Zicheng Liu, and Emad Barsoum.
\newblock Softvq-vae: Efficient 1-dimensional continuous tokenizer.
\newblock \emph{arXiv preprint arXiv:2412.10958}, 2024{\natexlab{a}}.

\bibitem[Chen et~al.(2025{\natexlab{b}})Chen, Han, Chen, Li, Wang, Wang, Wang,
  Liu, Zou, and Raj]{chen2025masked}
Hao Chen, Yujin Han, Fangyi Chen, Xiang Li, Yidong Wang, Jindong Wang, Ze Wang,
  Zicheng Liu, Difan Zou, and Bhiksha Raj.
\newblock Masked autoencoders are effective tokenizers for diffusion models.
\newblock \emph{arXiv preprint arXiv:2502.03444}, 2025{\natexlab{b}}.

\bibitem[Chen et~al.(2025{\natexlab{c}})Chen, Zhang, Tan, Guibas, Wetzstein,
  and Bi]{chen2025pi}
Hansheng Chen, Kai Zhang, Hao Tan, Leonidas Guibas, Gordon Wetzstein, and Sai
  Bi.
\newblock pi-flow: Policy-based few-step generation via imitation distillation.
\newblock \emph{arXiv preprint arXiv:2510.14974}, 2025{\natexlab{c}}.

\bibitem[Chen et~al.(2024{\natexlab{b}})Chen, Cai, Chen, Xie, Yang, Tang, Li,
  Lu, and Han]{chen2024deep}
Junyu Chen, Han Cai, Junsong Chen, Enze Xie, Shang Yang, Haotian Tang, Muyang
  Li, Yao Lu, and Song Han.
\newblock Deep compression autoencoder for efficient high-resolution diffusion
  models.
\newblock \emph{arXiv preprint arXiv:2410.10733}, 2024{\natexlab{b}}.

\bibitem[Chen et~al.(2024{\natexlab{c}})Chen, Ge, Xie, Wu, Yao, Ren, Wang, Luo,
  Lu, and Li]{chen2024pixart}
Junsong Chen, Chongjian Ge, Enze Xie, Yue Wu, Lewei Yao, Xiaozhe Ren, Zhongdao
  Wang, Ping Luo, Huchuan Lu, and Zhenguo Li.
\newblock Pixart-\$sigma\$: Weak-to-strong training of diffusion transformer
  for 4k text-to-image generation.
\newblock \emph{arXiv preprint arXiv:2403.04692}, 2024{\natexlab{c}}.

\bibitem[Chen et~al.(2024{\natexlab{d}})Chen, Jincheng, Chongjian, Yao, Xie,
  Wang, Kwok, Luo, Lu, and Li]{chenpixart}
Junsong Chen, YU Jincheng, GE Chongjian, Lewei Yao, Enze Xie, Zhongdao Wang,
  James Kwok, Ping Luo, Huchuan Lu, and Zhenguo Li.
\newblock Pixart-\$alpha\$: Fast training of diffusion transformer for
  photorealistic text-to-image synthesis.
\newblock In \emph{ICLR}, 2024{\natexlab{d}}.

\bibitem[Chen et~al.(2025{\natexlab{d}})Chen, Zou, He, Chen, Xie, Han, and
  Cai]{chen2025dc}
Junyu Chen, Dongyun Zou, Wenkun He, Junsong Chen, Enze Xie, Song Han, and Han
  Cai.
\newblock Dc-ae 1.5: Accelerating diffusion model convergence with structured
  latent space.
\newblock \emph{arXiv preprint arXiv:2508.00413}, 2025{\natexlab{d}}.

\bibitem[Dai et~al.(2023)Dai, Hou, Ma, Tsai, Wang, Wang, Zhang, Vandenhende,
  Wang, Dubey, et~al.]{dai2023emu}
Xiaoliang Dai, Ji Hou, Chih-Yao Ma, Sam Tsai, Jialiang Wang, Rui Wang, Peizhao
  Zhang, Simon Vandenhende, Xiaofang Wang, Abhimanyu Dubey, et~al.
\newblock Emu: Enhancing image generation models using photogenic needles in a
  haystack.
\newblock \emph{arXiv preprint arXiv:2309.15807}, 2023.

\bibitem[Esser et~al.(2024)Esser, Kulal, Blattmann, Entezari, M{\"u}ller,
  Saini, Levi, Lorenz, Sauer, Boesel, et~al.]{esser2024scaling}
Patrick Esser, Sumith Kulal, Andreas Blattmann, Rahim Entezari, Jonas
  M{\"u}ller, Harry Saini, Yam Levi, Dominik Lorenz, Axel Sauer, Frederic
  Boesel, et~al.
\newblock Scaling rectified flow transformers for high-resolution image
  synthesis.
\newblock In \emph{ICML}, 2024.

\bibitem[Fang et~al.(2024)Fang, Ma, and Wang]{fang2024structural}
Gongfan Fang, Xinyin Ma, and Xinchao Wang.
\newblock Structural pruning for diffusion models.
\newblock In \emph{NeurIPS}, 2024.

\bibitem[Frans et~al.(2024)Frans, Hafner, Levine, and Abbeel]{frans2024one}
Kevin Frans, Danijar Hafner, Sergey Levine, and Pieter Abbeel.
\newblock One step diffusion via shortcut models.
\newblock \emph{arXiv preprint arXiv:2410.12557}, 2024.

\bibitem[Geng et~al.(2025)Geng, Deng, Bai, Kolter, and He]{geng2025mean}
Zhengyang Geng, Mingyang Deng, Xingjian Bai, J~Zico Kolter, and Kaiming He.
\newblock Mean flows for one-step generative modeling.
\newblock \emph{arXiv preprint arXiv:2505.13447}, 2025.

\bibitem[Ghosh et~al.(2023)Ghosh, Hajishirzi, and Schmidt]{ghosh2023geneval}
Dhruba Ghosh, Hannaneh Hajishirzi, and Ludwig Schmidt.
\newblock Geneval: An object-focused framework for evaluating text-to-image
  alignment.
\newblock In \emph{NeurIPS}, 2023.

\bibitem[He et~al.(2025{\natexlab{a}})He, Gu, Chen, Zou, Lin, Zhang, Xi, Li,
  Zhu, Yu, et~al.]{he2025dc}
Wenkun He, Yuchao Gu, Junyu Chen, Dongyun Zou, Yujun Lin, Zhekai Zhang,
  Haocheng Xi, Muyang Li, Ligeng Zhu, Jincheng Yu, et~al.
\newblock Dc-gen: Post-training diffusion acceleration with deeply compressed
  latent space.
\newblock \emph{arXiv preprint arXiv:2509.25180}, 2025{\natexlab{a}}.

\bibitem[He et~al.(2025{\natexlab{b}})He, Gu, Chen, Zou, Lin, Zhang, Xi, Li,
  Zhu, Yu, et~al.]{he2025dcgen}
Wenkun He, Yuchao Gu, Junyu Chen, Dongyun Zou, Yujun Lin, Zhekai Zhang,
  Haocheng Xi, Muyang Li, Ligeng Zhu, Jincheng Yu, et~al.
\newblock Dc-gen: Post-training diffusion acceleration with deeply compressed
  latent space.
\newblock \emph{arXiv preprint arXiv:2509.25180}, 2025{\natexlab{b}}.

\bibitem[Heusel et~al.(2017)Heusel, Ramsauer, Unterthiner, Nessler, and
  Hochreiter]{fid}
Martin Heusel, Hubert Ramsauer, Thomas Unterthiner, Bernhard Nessler, and Sepp
  Hochreiter.
\newblock Gans trained by a two time-scale update rule converge to a local nash
  equilibrium.
\newblock In \emph{NeurIPS}, 2017.

\bibitem[Ho et~al.(2020)Ho, Jain, and Abbeel]{ho2020denoising}
Jonathan Ho, Ajay Jain, and Pieter Abbeel.
\newblock Denoising diffusion probabilistic models.
\newblock In \emph{NeurIPS}, 2020.

\bibitem[Kim et~al.(2025)Kim, He, Yu, Yang, Shen, Kwak, and
  Chen]{kim2025democratizing}
Dongwon Kim, Ju He, Qihang Yu, Chenglin Yang, Xiaohui Shen, Suha Kwak, and
  Liang-Chieh Chen.
\newblock Democratizing text-to-image masked generative models with compact
  text-aware one-dimensional tokens.
\newblock \emph{arXiv preprint arXiv:2501.07730}, 2025.

\bibitem[Kirillov et~al.(2023)Kirillov, Mintun, Ravi, Mao, Rolland, Gustafson,
  Xiao, Whitehead, Berg, Lo, et~al.]{kirillov2023segment}
Alexander Kirillov, Eric Mintun, Nikhila Ravi, Hanzi Mao, Chloe Rolland, Laura
  Gustafson, Tete Xiao, Spencer Whitehead, Alexander~C Berg, Wan-Yen Lo, et~al.
\newblock Segment anything.
\newblock In \emph{ICCV}, 2023.

\bibitem[Labs(2024)]{flux2024}
Black~Forest Labs.
\newblock Flux.
\newblock \emph{Online}, 2024.

\bibitem[Li et~al.(2024{\natexlab{a}})Li, Kamko, Akhgari, Sabet, Xu, and
  Doshi]{li2024playground}
Daiqing Li, Aleks Kamko, Ehsan Akhgari, Ali Sabet, Linmiao Xu, and Suhail
  Doshi.
\newblock Playground v2. 5: Three insights towards enhancing aesthetic quality
  in text-to-image generation.
\newblock \emph{arXiv preprint arXiv:2402.17245}, 2024{\natexlab{a}}.

\bibitem[Li et~al.(2024{\natexlab{b}})Li, Lin, Zhang, Cai, Li, Guo, Xie, Meng,
  Zhu, and Han]{li2024svdquant}
Muyang Li, Yujun Lin, Zhekai Zhang, Tianle Cai, Xiuyu Li, Junxian Guo, Enze
  Xie, Chenlin Meng, Jun-Yan Zhu, and Song Han.
\newblock Svdquant: Absorbing outliers by low-rank components for 4-bit
  diffusion models.
\newblock \emph{arXiv preprint arXiv:2411.05007}, 2024{\natexlab{b}}.

\bibitem[Luo et~al.(2023)Luo, Tan, Huang, Li, and Zhao]{luo2023latent}
Simian Luo, Yiqin Tan, Longbo Huang, Jian Li, and Hang Zhao.
\newblock Latent consistency models: Synthesizing high-resolution images with
  few-step inference.
\newblock \emph{arXiv preprint arXiv:2310.04378}, 2023.

\bibitem[Ma et~al.(2024)Ma, Fang, and Wang]{ma2024deepcache}
Xinyin Ma, Gongfan Fang, and Xinchao Wang.
\newblock Deepcache: Accelerating diffusion models for free.
\newblock In \emph{CVPR}, 2024.

\bibitem[Meng et~al.(2023)Meng, Rombach, Gao, Kingma, Ermon, Ho, and
  Salimans]{meng2023distillation}
Chenlin Meng, Robin Rombach, Ruiqi Gao, Diederik Kingma, Stefano Ermon,
  Jonathan Ho, and Tim Salimans.
\newblock On distillation of guided diffusion models.
\newblock In \emph{CVPR}, 2023.

\bibitem[Peebles and Xie(2023)]{peebles2023scalable}
William Peebles and Saining Xie.
\newblock Scalable diffusion models with transformers.
\newblock In \emph{ICCV}, 2023.

\bibitem[Peng et~al.(2025)Peng, Zheng, Shen, Young, Guo, Wang, Xu, Liu, Jiang,
  Li, et~al.]{peng2025open}
Xiangyu Peng, Zangwei Zheng, Chenhui Shen, Tom Young, Xinying Guo, Binluo Wang,
  Hang Xu, Hongxin Liu, Mingyan Jiang, Wenjun Li, et~al.
\newblock Open-sora 2.0: Training a commercial-level video generation model in
  \$200 k.
\newblock \emph{arXiv preprint arXiv:2503.09642}, 2025.

\bibitem[Podell et~al.(2023)Podell, English, Lacey, Blattmann, Dockhorn,
  M{\"u}ller, Penna, and Rombach]{podell2023sdxl}
Dustin Podell, Zion English, Kyle Lacey, Andreas Blattmann, Tim Dockhorn, Jonas
  M{\"u}ller, Joe Penna, and Robin Rombach.
\newblock Sdxl: Improving latent diffusion models for high-resolution image
  synthesis.
\newblock \emph{arXiv preprint arXiv:2307.01952}, 2023.

\bibitem[Rombach et~al.(2022)Rombach, Blattmann, Lorenz, Esser, and
  Ommer]{rombach2022high}
Robin Rombach, Andreas Blattmann, Dominik Lorenz, Patrick Esser, and Bj{\"o}rn
  Ommer.
\newblock High-resolution image synthesis with latent diffusion models.
\newblock In \emph{CVPR}, 2022.

\bibitem[Saharia et~al.(2022)Saharia, Chan, Saxena, Li, Whang, Denton,
  Ghasemipour, Gontijo~Lopes, Karagol~Ayan, Salimans,
  et~al.]{saharia2022photorealistic}
Chitwan Saharia, William Chan, Saurabh Saxena, Lala Li, Jay Whang, Emily~L
  Denton, Kamyar Ghasemipour, Raphael Gontijo~Lopes, Burcu Karagol~Ayan, Tim
  Salimans, et~al.
\newblock Photorealistic text-to-image diffusion models with deep language
  understanding.
\newblock In \emph{NeurIPS}, 2022.

\bibitem[Salimans and Ho(2022)]{salimans2022progressive}
Tim Salimans and Jonathan Ho.
\newblock Progressive distillation for fast sampling of diffusion models.
\newblock In \emph{ICLR}, 2022.

\bibitem[Shi et~al.(2025)Shi, Wang, Zheng, Yuan, Wu, Wang, Wan, Zhou, and
  Lu]{shi2025latent}
Minglei Shi, Haolin Wang, Wenzhao Zheng, Ziyang Yuan, Xiaoshi Wu, Xintao Wang,
  Pengfei Wan, Jie Zhou, and Jiwen Lu.
\newblock Latent diffusion model without variational autoencoder.
\newblock \emph{arXiv preprint arXiv:2510.15301}, 2025.

\bibitem[Shih et~al.(2024)Shih, Belkhale, Ermon, Sadigh, and
  Anari]{shih2024parallel}
Andy Shih, Suneel Belkhale, Stefano Ermon, Dorsa Sadigh, and Nima Anari.
\newblock Parallel sampling of diffusion models.
\newblock In \emph{NeurIPS}, 2024.

\bibitem[{Stability AI}(2024)]{sd35}
{Stability AI}.
\newblock Sd3.5.
\newblock \url{https://github.com/Stability-AI/sd3.5}, 2024.

\bibitem[Tang et~al.(2024)Tang, Tang, Luo, Wang, and
  Chang]{tang2024accelerating}
Zhiwei Tang, Jiasheng Tang, Hao Luo, Fan Wang, and Tsung-Hui Chang.
\newblock Accelerating parallel sampling of diffusion models.
\newblock In \emph{ICML}, 2024.

\bibitem[Wang et~al.(2024)Wang, Fang, Li, and Yang]{wang2024pipefusion}
Jiannan Wang, Jiarui Fang, Aoyu Li, and PengCheng Yang.
\newblock Pipefusion: Displaced patch pipeline parallelism for inference of
  diffusion transformer models.
\newblock \emph{arXiv preprint arXiv:2405.14430}, 2024.

\bibitem[Wang et~al.(2004)Wang, Bovik, Sheikh, and Simoncelli]{ssim}
Zhou Wang, Alan~C Bovik, Hamid~R Sheikh, and Eero~P Simoncelli.
\newblock Image quality assessment: from error visibility to structural
  similarity.
\newblock \emph{IEEE TIP}, 2004.

\bibitem[Wang et~al.(2023)Wang, Montoya, Munechika, Yang, Hoover, and
  Chau]{wang2023diffusiondb}
Zijie~J Wang, Evan Montoya, David Munechika, Haoyang Yang, Benjamin Hoover, and
  Duen~Horng Chau.
\newblock Diffusiondb: A large-scale prompt gallery dataset for text-to-image
  generative models.
\newblock In \emph{ACL}, 2023.

\bibitem[Xie et~al.(2024)Xie, Chen, Chen, Cai, Tang, Lin, Zhang, Li, Zhu, Lu,
  et~al.]{xie2024sana}
Enze Xie, Junsong Chen, Junyu Chen, Han Cai, Haotian Tang, Yujun Lin, Zhekai
  Zhang, Muyang Li, Ligeng Zhu, Yao Lu, et~al.
\newblock Sana: Efficient high-resolution image synthesis with linear diffusion
  transformers.
\newblock \emph{arXiv preprint arXiv:2410.10629}, 2024.

\bibitem[Xie et~al.(2025{\natexlab{a}})Xie, Chen, Zhao, Yu, Zhu, Lin, Zhang,
  Li, Chen, Cai, et~al.]{xie2025sana}
Enze Xie, Junsong Chen, Yuyang Zhao, Jincheng Yu, Ligeng Zhu, Yujun Lin, Zhekai
  Zhang, Muyang Li, Junyu Chen, Han Cai, et~al.
\newblock Sana 1.5: Efficient scaling of training-time and inference-time
  compute in linear diffusion transformer.
\newblock \emph{arXiv preprint arXiv:2501.18427}, 2025{\natexlab{a}}.

\bibitem[Xie et~al.(2025{\natexlab{b}})Xie, Zhang, Huang, Zhang, Lu, and
  Yang]{xie2025layton}
Qingsong Xie, Zhao Zhang, Zhe Huang, Yanhao Zhang, Haonan Lu, and Zhenyu Yang.
\newblock Layton: Latent consistency tokenizer for 1024-pixel image
  reconstruction and generation by 256 tokens.
\newblock \emph{arXiv preprint arXiv:2503.08377}, 2025{\natexlab{b}}.

\bibitem[Yao et~al.(2025)Yao, Yang, and Wang]{yao2025vavae}
Jingfeng Yao, Bin Yang, and Xinggang Wang.
\newblock {Reconstruction vs. Generation}: Taming optimization dilemma in
  latent diffusion models.
\newblock In \emph{CVPR}, 2025.

\bibitem[Yin et~al.(2024{\natexlab{a}})Yin, Gharbi, Park, Zhang, Shechtman,
  Durand, and Freeman]{yin2024improved}
Tianwei Yin, Micha{\"e}l Gharbi, Taesung Park, Richard Zhang, Eli Shechtman,
  Fredo Durand, and William~T Freeman.
\newblock Improved distribution matching distillation for fast image synthesis.
\newblock \emph{arXiv preprint arXiv:2405.14867}, 2024{\natexlab{a}}.

\bibitem[Yin et~al.(2024{\natexlab{b}})Yin, Gharbi, Zhang, Shechtman, Durand,
  Freeman, and Park]{yin2024one}
Tianwei Yin, Micha{\"e}l Gharbi, Richard Zhang, Eli Shechtman, Fredo Durand,
  William~T Freeman, and Taesung Park.
\newblock One-step diffusion with distribution matching distillation.
\newblock In \emph{CVPR}, 2024{\natexlab{b}}.

\bibitem[Yin et~al.(2024{\natexlab{c}})Yin, Gharbi, Zhang, Shechtman, Durand,
  Freeman, and Park]{yin2024onestep}
Tianwei Yin, Micha{\"e}l Gharbi, Richard Zhang, Eli Shechtman, Fr{\'e}do
  Durand, William~T Freeman, and Taesung Park.
\newblock One-step diffusion with distribution matching distillation.
\newblock In \emph{CVPR}, 2024{\natexlab{c}}.

\bibitem[Yu et~al.(2024{\natexlab{a}})Yu, Weber, Deng, Shen, Cremers, and
  Chen]{yu2024image}
Qihang Yu, Mark Weber, Xueqing Deng, Xiaohui Shen, Daniel Cremers, and
  Liang-Chieh Chen.
\newblock An image is worth 32 tokens for reconstruction and generation.
\newblock In \emph{NeurIPS}, 2024{\natexlab{a}}.

\bibitem[Yu et~al.(2024{\natexlab{b}})Yu, Kwak, Jang, Jeong, Huang, Shin, and
  Xie]{yu2024representation}
Sihyun Yu, Sangkyung Kwak, Huiwon Jang, Jongheon Jeong, Jonathan Huang, Jinwoo
  Shin, and Saining Xie.
\newblock Representation alignment for generation: Training diffusion
  transformers is easier than you think.
\newblock \emph{arXiv preprint arXiv:2410.06940}, 2024{\natexlab{b}}.

\bibitem[Yue et~al.(2025)Yue, Zhang, Zeng, Chen, Wang, Zhuang, Dong, Du, Wang,
  Wang, et~al.]{yue2025uniflow}
Zhengrong Yue, Haiyu Zhang, Xiangyu Zeng, Boyu Chen, Chenting Wang, Shaobin
  Zhuang, Lu Dong, KunPeng Du, Yi Wang, Limin Wang, et~al.
\newblock Uniflow: A unified pixel flow tokenizer for visual understanding and
  generation.
\newblock \emph{arXiv preprint arXiv:2510.10575}, 2025.

\bibitem[Zhang et~al.(2025)Zhang, Huang, Liu, Guo, and
  Huang]{zhang2025diffusion}
Jinjin Zhang, Qiuyu Huang, Junjie Liu, Xiefan Guo, and Di Huang.
\newblock Diffusion-4k: Ultra-high-resolution image synthesis with latent
  diffusion models.
\newblock In \emph{CVPR}, 2025.

\bibitem[Zhang and Chen(2023)]{zhangfast}
Qinsheng Zhang and Yongxin Chen.
\newblock Fast sampling of diffusion models with exponential integrator.
\newblock In \emph{ICLR}, 2023.

\bibitem[Zhang et~al.(2023)Zhang, Tao, and Chen]{zhanggddim}
Qinsheng Zhang, Molei Tao, and Yongxin Chen.
\newblock gddim: Generalized denoising diffusion implicit models.
\newblock In \emph{ICLR}, 2023.

\bibitem[Zhang et~al.(2018)Zhang, Isola, Efros, Shechtman, and
  Wang]{zhang2018unreasonable}
Richard Zhang, Phillip Isola, Alexei~A Efros, Eli Shechtman, and Oliver Wang.
\newblock The unreasonable effectiveness of deep features as a perceptual
  metric.
\newblock In \emph{CVPR}, 2018.

\bibitem[Zhao et~al.(2024{\natexlab{a}})Zhao, Fang, Liu, Rui, Soedarmadji, Li,
  Lin, Dai, Yan, Yang, et~al.]{zhao2024vidit}
Tianchen Zhao, Tongcheng Fang, Enshu Liu, Wan Rui, Widyadewi Soedarmadji,
  Shiyao Li, Zinan Lin, Guohao Dai, Shengen Yan, Huazhong Yang, et~al.
\newblock Vidit-q: Efficient and accurate quantization of diffusion
  transformers for image and video generation.
\newblock \emph{arXiv preprint arXiv:2406.02540}, 2024{\natexlab{a}}.

\bibitem[Zhao et~al.(2024{\natexlab{b}})Zhao, Bai, Rao, Zhou, and
  Lu]{zhao2024unipc}
Wenliang Zhao, Lujia Bai, Yongming Rao, Jie Zhou, and Jiwen Lu.
\newblock Unipc: A unified predictor-corrector framework for fast sampling of
  diffusion models.
\newblock In \emph{NeurIPS}, 2024{\natexlab{b}}.

\bibitem[Zheng et~al.(2025)Zheng, Ma, Tong, and Xie]{zheng2025diffusion}
Boyang Zheng, Nanye Ma, Shengbang Tong, and Saining Xie.
\newblock Diffusion transformers with representation autoencoders.
\newblock \emph{arXiv preprint arXiv:2510.11690}, 2025.

\bibitem[Zheng et~al.(2023)Zheng, Lu, Chen, and Zhu]{zheng2023dpm}
Kaiwen Zheng, Cheng Lu, Jianfei Chen, and Jun Zhu.
\newblock Dpm-solver-v3: Improved diffusion ode solver with empirical model
  statistics.
\newblock In \emph{NeurIPS}, 2023.

\bibitem[Zhengwentai(2023)]{taited2023CLIPScore}
SUN Zhengwentai.
\newblock {clip-score: CLIP Score for PyTorch}.
\newblock \url{https://github.com/taited/clip-score}, 2023.
\newblock Version 0.2.1.

\bibitem[Zhou et~al.(2025)Zhou, Ermon, and Song]{zhou2025inductive}
Linqi Zhou, Stefano Ermon, and Jiaming Song.
\newblock Inductive moment matching.
\newblock \emph{arXiv preprint arXiv:2503.07565}, 2025.

\bibitem[Zhu et~al.(2023)Zhu, Feng, Chen, Bao, Wang, Chen, Yuan, and
  Hua]{zhu2023designing}
Zixin Zhu, Xuelu Feng, Dongdong Chen, Jianmin Bao, Le Wang, Yinpeng Chen, Lu
  Yuan, and Gang Hua.
\newblock Designing a better asymmetric vqgan for stablediffusion.
\newblock \emph{arXiv preprint arXiv:2306.04632}, 2023.

\end{thebibliography}
